\documentclass[journal,comsoc]{IEEEtran}
\pdfoutput=1
\usepackage[T1]{fontenc}
\usepackage{makecell}

\usepackage{graphicx}  
\usepackage{float}  
\usepackage{subfigure}  
\usepackage{booktabs}
\usepackage{multirow}
\usepackage{amssymb}
\usepackage{color}
\usepackage{amsmath}
\usepackage{algorithm2e}
\RestyleAlgo{ruled}
\usepackage{enumitem}
\usepackage[pagebackref=true, colorlinks]{hyperref}
\usepackage{lipsum}

\newlength{\bibitemsep}
\newlength{\bibparskip}
\let\oldthebibliography\thebibliography
\renewcommand\thebibliography[1]{%
  \oldthebibliography{#1}%
  \setlength{\parskip}{\bibitemsep}%
  \setlength{\itemsep}{\bibparskip}%
}

\usepackage[cmintegrals]{newtxmath}
\newcommand{\etal}{\textit{et~al.}}

\begin{document}

\title{Learning Cross-view Visual Geo-localization without Ground Truth}

\author {
Haoyuan~Li,
\and Chang~Xu, 
\and Wen~Yang, 
\and Huai~Yu,~
\and Gui-Song~Xia

\thanks{This work was supported in part by the National Natural Science Foundation of China (NSFC) under Grant 62271355 and NSFC Regional Innovation and Development Joint Fund (No. U22A2010).}
\thanks{H. Li, C. Xu, W. Yang, and H. Yu are with the School of Electronic Information, Wuhan University, Wuhan, 430072 China. \emph{E-mail: \{lihaoyuan, xuchangeis, yangwen, yuhuai\}@whu.edu.cn}}
\thanks{G-S. Xia is with the School of Computer Science and the State Key Lab. LIESMARS, Wuhan University, Wuhan, 430072, China. \emph{E-mail: guisong.xia@whu.edu.cn}}
}

\maketitle

\begin{abstract}

Cross-View Geo-Localization (CVGL) involves determining the geographical location of a query image by matching it with a corresponding GPS-tagged reference image. Current state-of-the-art methods predominantly rely on training models with labeled paired images, incurring substantial annotation costs and training burdens. In this study, we investigate the adaptation of frozen models for CVGL without requiring ground truth pair labels. We observe that training on unlabeled cross-view images presents significant challenges, including the need to establish relationships within unlabeled data and reconcile view discrepancies between uncertain queries and references. To address these challenges, we propose a self-supervised learning framework to train a learnable adapter for a frozen Foundation Model (FM). This adapter is designed to map feature distributions from diverse views into a uniform space using unlabeled data exclusively. To establish relationships within unlabeled data, we introduce an Expectation-Maximization-based Pseudo-labeling module, which iteratively estimates associations between cross-view features and optimizes the adapter. To maintain the robustness of the FM's representation, we incorporate an information consistency module with a reconstruction loss, ensuring that adapted features retain strong discriminative ability across views. Experimental results demonstrate that our proposed method achieves significant improvements over vanilla FMs and competitive accuracy compared to supervised methods, while necessitating fewer training parameters and relying solely on unlabeled data. Evaluation of our adaptation for task-specific models further highlights its broad applicability. Particularly, on the University-1652 dataset, our method outperforms the FM baseline by a substantial margin, achieving about 39 points improvement in Recall@1 and more than 34 points increase in Average Precision. Codes will be released soon.

\end{abstract}

\begin{IEEEkeywords}
Cross-view Geo-localization, Self-supervised Learning, Foundation Model
\end{IEEEkeywords}

\IEEEpeerreviewmaketitle

\section{Introduction}

\begin{figure}
    \centering
    \includegraphics[width=0.49\textwidth]{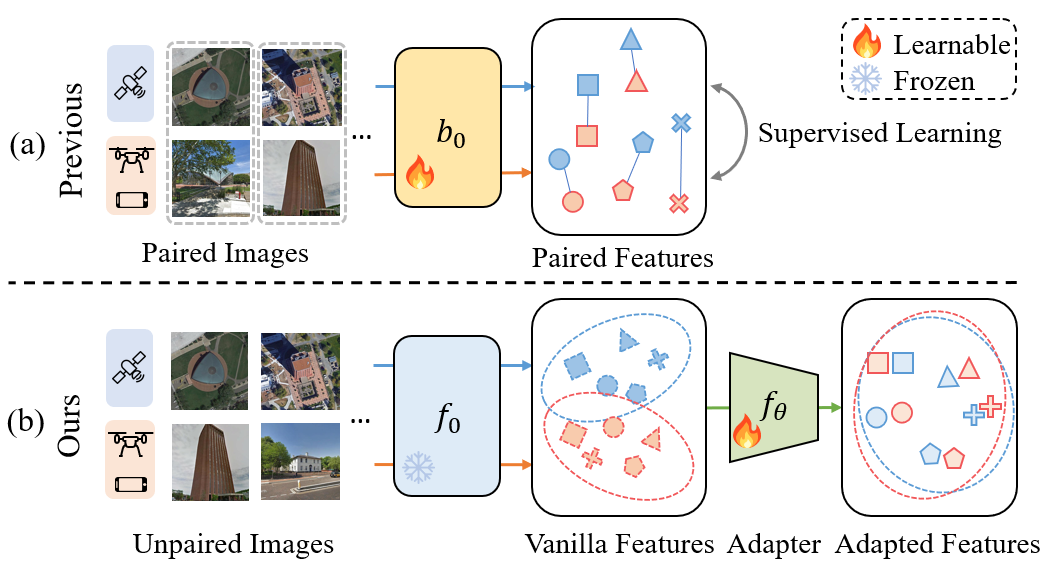}
    \centering
    \caption{\textbf{Illustration of the previous supervised paradigm (a) and the proposed self-supervised paradigm (b).} $b_0$ denotes backbone, $f_0$ denotes the frozen model, and $f_\theta$ denotes the learnable adapter.}
    \label{fig:challenge}
\end{figure}

\IEEEPARstart{C}{ross-view} Geo-Localization (CVGL) aims at determining the geo-location of a query image by retrieving its corresponding geo-tagged overhead reference image~\cite{ren2023hashing}. Benefitting application scenarios like navigation~\cite{xue2022terrain}, urban planning, and building localization, cross-view geo-localization is mainly addressed with a fully supervised paradigm in the existing literatures~\cite{lu2023okay, sun2023cross}. As illustrated in Fig.~\ref{fig:challenge}(a), a ground-view query image along with a paired overhead reference image is required to supervise the feature embedding alignment.

Nevertheless, these methods are constrained by the availability of location information during query image capturing for data annotation. For example, when the GPS signal is dispersed or absent in scenarios like crowded cities and urban canyons, the availability of precise location information is not guaranteed. In drone navigation, the tilted shooting angle of the drone introduces a positional offset between the drone's location and the captured scene, which hinders the direct use of the drone's GPS as a location label for the image. Besides, images on the website usually lack location information or precise geo-tags, making them unsuitable for supervising existing pipelines. This dependency on annotation imposes limitations on supervised methods in real-world scenarios. On the other hand, these methods typically necessitate the parameter updating of the entire model, leading to heavy computation and time costs. To address the two above limitations of the existing procedures, we aim to diminish the reliance on labeled data and release the computation burden of updating the entire model. Considering the robust zero-shot capability of feature representation in 
recent large-scale pre-trained models, commonly referred to as Foundation Models (FMs), an effective way is to employ the frozen FM on the geo-localization task, which is illustrated in Fig.~\ref{fig:challenge}(b). 


\begin{figure}
    \centering
    \includegraphics[width=0.45\textwidth]{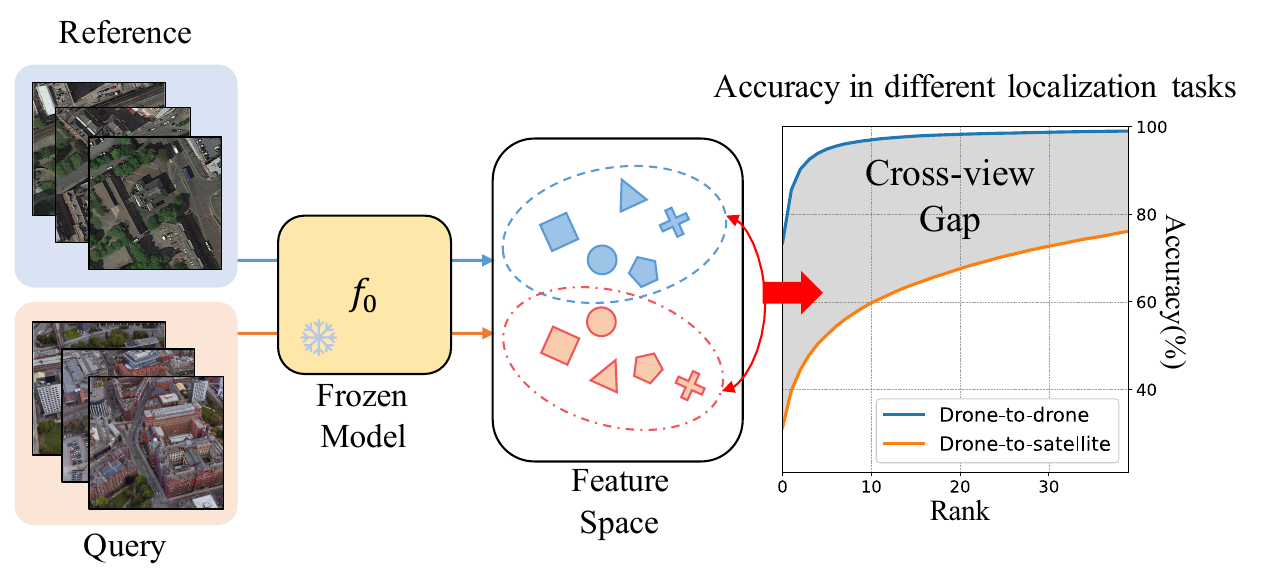}
    \centering
    \caption{\textbf{Performance degradation of the foundation model due to the view gap.} We perform Drone-to-drone (single-view) and Drone-to-satellite (cross-view) retrieval using the frozen foundation model. The blue curve represents the retrieval accuracy for single-view, while the red curve illustrates the lower accuracy for cross-view. More experimental details are presented in Section~\ref{sec:ablation}.}
    \label{fig:view-gap}
\end{figure}

While foundation models have been well-adapted for specialized tasks such as classification and segmentation tasks, existing approaches either rely on labeled data for training adapters or are limited to single-domain target tasks. In contrast, CVGL involves the adaptation to two distinct target domains, which presents a unique challenge. Addressing this challenge requires not only learning suitable feature representations but also bridging the domain gap inherent in the cross-view scenario itself.
To investigate foundation models' generalization across different views, we conduct experiments on the University-1652 dataset~\cite{zheng2020university} as shown in Fig.~\ref{fig:view-gap}. The foundation model exhibits satisfactory performance when handling the single-view geo-localization task, achieving considerable accuracy in the Drone-to-drone scenario (blue curve on the right figure). However, the accuracy drops significantly when it comes to Drone-to-satellite cross-view scenarios (red curve on the right figure). 
This inspires us to adapt the frozen foundation model's output to mitigate the cross-view feature discrepancy. Considering the absence of pair labels for images, the two following challenges need to be addressed when adapting the foundation model to the CVGL.

\begin{enumerate}
    \item \textbf{Mining relationships from unlabeled data.} 
    In previous works, matched image pairs are identified using annotation labels, and models are trained with the ground truth's supervision. However, in the absence of ground truth relationships, the challenge lies in identifying the potential positive targets from unlabeled data, which becomes even more difficult when images exhibit cross-view variance. 

    \item \textbf{Preserving Robustness of Frozen Models.}
    Adaptation can help the frozen models to better suit new scenarios. However, the significant view discrepancy and lack of ground truth may cause the model to overfit the training data, harming the original feature representation's robustness. Effectively adapting the frozen model for the new cross-view tasks while preserving the original feature representation's robustness poses a significant challenge.
    
\end{enumerate}

To tackle these challenges, we introduce a self-supervised adaptation pipeline tailored for cross-view geo-localization, aiming to bridge the gap between diverse view perspectives. Initially, we leverage a frozen foundation model to extract initial features from cross-view images, each exhibiting distinct distributions. Subsequently, we propose an adapter to harmonize these initial features into a uniform representation. To optimize the adapter, we devise an Expectation Maximization learning pipeline, extracting potential positive samples from unlabeled data. Furthermore, to maintain the robustness of the foundation model, we introduce an Adaptation Information Consistency (AIC) module, ensuring consistency between initial and adapted features. Extensive experiments validate the efficacy of our pipeline in adapting the foundation model for Drone-to-satellite geo-localization, achieving performance on par with supervised methods. Notably, our approach demonstrates promising improvements even in the more challenging Ground-to-satellite scenario, enhancing the performance of previously well-trained models on unseen datasets across different cities.

In summary, our contributions are three-fold:
\begin{itemize}	
\item We propose a self-supervised adaptation pipeline for cross-view geo-localization without ground truth, leveraging a frozen foundation model to extract initial features and an adapter to unify them into a consistent representation. 
\item We introduce an EM-based Pseudo-Labeling (EMPL) module to estimate positive samples from unlabeled data and optimize the adapter, alongside an Adaptation Information Consistency (AIC) module to ensure feature consistency.

\item We validate the effectiveness of our approach through comprehensive experiments, showcasing its capability to adapt the foundation model for Drone-to-satellite geo-localization while achieving comparable performance to supervised methods. Furthermore, our method can enhance the performance of task-specific pre-trained models on new datasets across cities, even in the more demanding Ground-to-satellite geo-localization scenario.
\end{itemize}

The rest of the paper is structured as follows: Section~\ref{sec:related_work} offers a succinct survey of the related works. In Section~\ref{sec:method}, we provide a comprehensive exposition of our proposed pipeline, including the overview of the framework, and the self-supervised learning paradigm for the adapter. Section~\ref{sec:experiment} presents and analyzes the experimental results. Section~\ref{sec:discussion} discusses the limitations of the proposed scheme, as well as possible avenues for future research. Finally, we conclude in Section~\ref{sec:conclusion}.

\section{Related Work}
\label{sec:related_work}
In this section, we will review the recent progress in cross-view geo-localization, the development of the foundation models on geo-localization, and the self-supervised adaptation.

\subsection{Cross-view Geo-localization}
The Cross-view Geo-localization (CVGL) task involves performing image retrieval to find relevant images from a reference database of images given a query image, where the query and reference image are from different perspectives or sensors. In this field of CVGL, various works have been developed to address the challenges posed by tasks such as Drone-to-satellite and ground-to-satellite localization. 

For the Drone-to-satellite geo-localization, Zheng~\etal~\cite{zheng2020university} form the first multi-view multi-source dataset for this task and employ classification loss for discriminating scenes. RK-Net~\cite{lin2022joint} focuses on common salient point detection for cross-view image pairs. LPN~\cite{wang2021each} suggests the neighbor areas can provide auxiliary information and introduce a feature partitioning strategy to enhance feature representation. Considering the limitation of the recent CNN, some works~\cite{dai2021transformer, zhao2024transfg} design Transformer-based networks to tackle global information of the images and address the scale discrepancy.

For ground-to-satellite geo-localization, popular datasets CVUSA~\cite{workman2015wide} and CVACT~\cite{liu2019lending} provide the benchmarks. These datasets consist of one-to-one drastic cross-view image pairs with center location alignment, ensuring the north direction of panoramic photos aligns with the center top of the satellite images. CVM-Net~\cite{hu2018cvm} first employs two weight-shared NetVLADs~\cite{arandjelovic2016netvlad} to embed images into a common space and obtain global descriptors to address the view discrepancy. Regmi~\etal~\cite{regmi2019bridging} leverage GANs for view synthesis of aerial images to eliminate view discrepancy from the image appearance. Considering the fixed direction alignment between queries and references may make the network overfit to the training set, GeoDTR~\cite{zhang2023cross} explicitly disentangles geometric information from raw features and mitigates overfitting on low-level details. While hard negative selection is important for the overall performance, Sample4Geo~\cite{Deuser_2023_ICCV} proposes two sampling strategies for mining hard negatives, including the geographical neighbors and visual similarity in image embeddings. 

While recent methodologies have made significant contributions, their training pipelines rely heavily on annotated data, often requiring information such as scene identification or labeled image pairs. Moreover, the prevalent training pipelines necessitate the training of the entire model, incurring computational expenses and demanding substantial time resources. In response to these limitations, we introduce a self-supervised learning pipeline on unlabeled data for cross-view geo-localization.

\subsection{Foundation Model in Geo-localization}

With the advent of Foundation Models (FMs), recent advancements in cross-view geo-localization have incorporated them as the initial backbones of the training process. Prominent examples of popular self-supervised visual encoder backbones include DINO and DINOv2~\cite{caron2021emerging, oquab2024dinov}. The former is pre-trained on ImageNet~\cite{deng2009imagenet}, while the latter is pre-trained on a larger dataset. CLIP~\cite{radford2021learning}, renowned for its ability in text and visual contrastive learning, leverages its robust language capabilities for multi-modal retrieval tasks. SAM~\cite{kirillov2023segment} trains the segmentation network in its large-scale image dataset, which achieves zero-shot generalization in various segmentation tasks.

In cross-view geo-localization, research has explored diverse approaches for applying the foundation models. IM2City~\cite{wu2022im2city} incorporates natural language information from image labels to enhance embedding corresponding visual features. GeoCLIP~\cite{cepeda2023geoclip} takes a different route by embedding GPS locations through the CLIP text encoder, facilitating alignment between images and their geographical coordinates. StreetCLIP~\cite{haas2023learning} introduces a meta-learning framework for the visual feature encoder of the original CLIP, which aims to improve the model's zero-shot capability in addressing open-domain geo-localization. AnyLoc~\cite{keetha2023anyloc} is a universal technique that generalizes the foundation model across various visual localization tasks through manually designed feature aggregation. 

While the direct generalization of the aforementioned foundation models demonstrates promising performance in various geo-localization tasks~\cite {keetha2023anyloc}, they encounter degradation in performance due to the cross-view distribution discrepancy. To overcome this challenge, we delve into adaptation to mitigate the view discrepancy and enhance the performance of the foundation models. 

\subsection{Self-supervised Adaptation}

Many works~\cite{xiao2023visual, pantazis2022svl, pan2023self, lu2023towards} focus on adapting the foundation model and enhancing the performance in the downstream task. SVL-Adapter~\cite{pantazis2022svl} is a self-supervised adapter for CLIP to improve the performance on various downstream visual classification tasks. For the segmentation downstream task, SAM-adapter~\cite{chen2023sam} enhances SAM by incorporating domain-specific information through the simple adapters. SUPMER~\cite{pan2023self} employs self-supervised meta-learning with a diverse set of meta-training tasks to acquire a universal prompt initialization for adaptation, relying solely on unlabeled data. Yang~\etal~\cite{yang2020mining} focus on bolstering zero-shot performance during testing by mining unseen classes, still requiring the labeled data.

Other endeavors aim to mitigate the burden associated with supervised learning. Xu~\etal~\cite{xu2022bayesian} establish a connection between pseudo-labeling and the Expectation Maximization algorithm (EM)~\cite{neal1998view}, providing an empirical explanation for its success. Sumbul~\etal~\cite{sumbul2022novel} tackle the annotation challenges in remote sensing retrieval, devising a self-supervised learning pipeline. Acknowledging the complexities of real-world scenarios of geo-localization, Wang~\etal~\cite{wang2022multiple} synthesize multiple environmental styles and propose a self-adaptive network to mitigate domain shifts induced by environmental changes.

The adaptation above has primarily concentrated on adjusting models to single-domain target datasets. However, in cross-view geo-localization, the target domains inherently involve two distinct domains due to view discrepancies. Consequently, our adaptation process must not only focus on adapting the model to unseen data in the downstream CVGL task but also address the challenge posed by the discrepancy between cross-view features within these target domains.

\section{Methodology}
\label{sec:method}
\begin{figure*}
	\centering
	\includegraphics[width=0.9\textwidth]{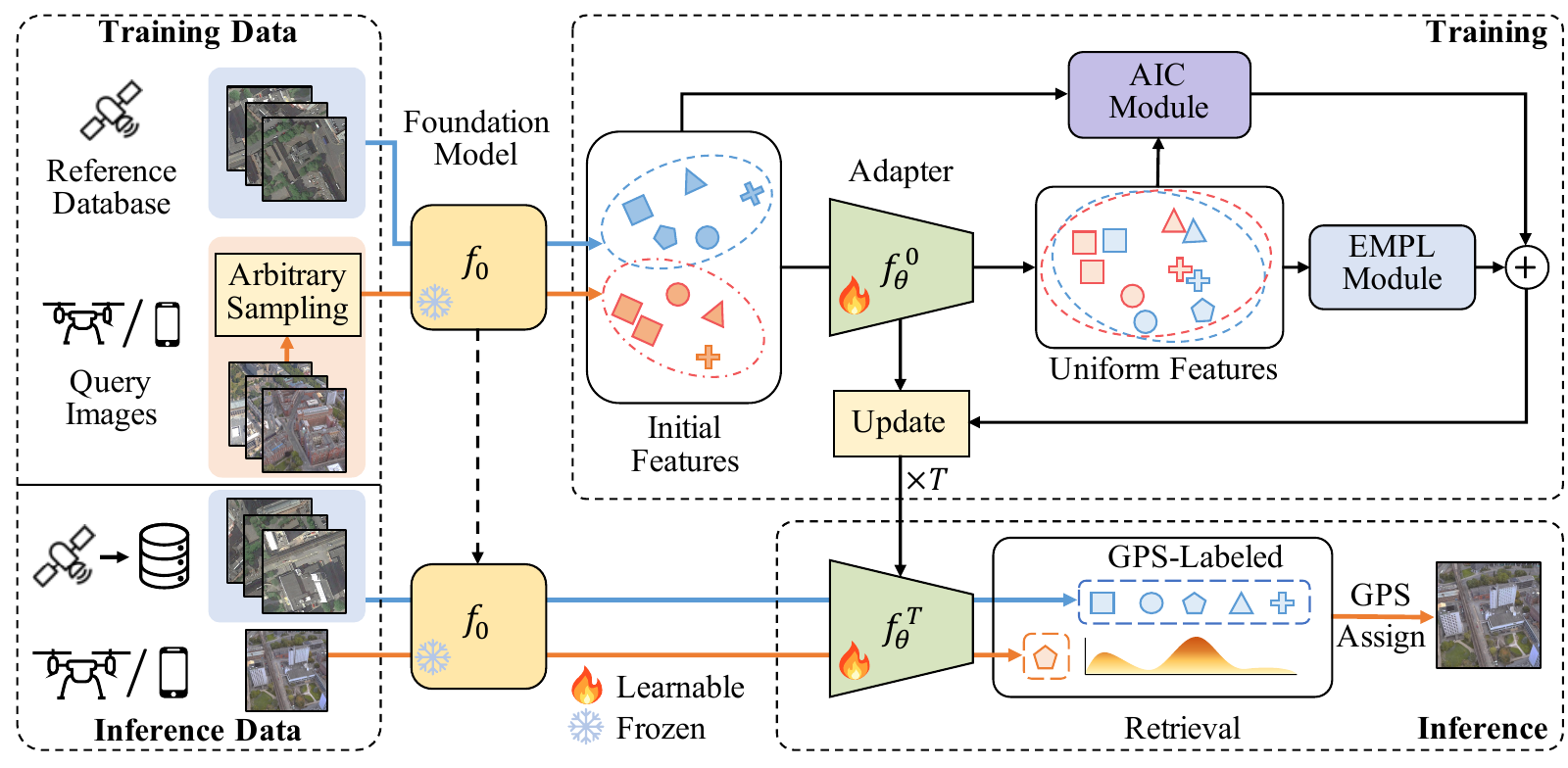}
	\caption{\textbf{Overview of the self-supervised cross-view adaptation.} In the training phase, the foundation model is frozen and the adapter is trained via the proposed EMPL and AIC modules without ground truth. In the inference phase, the global features of input images are extracted by the frozen foundation modal and the trained adapter for retrieval to final geo-localization. }
	\label{fig:emb_frame}
\end{figure*}

\subsection{Problem Definition and Notations}

For cross-view geo-localization, the query image set $\mathcal{I}^Q=\{\mathbf{I}^Q_1, ..., \mathbf{I}^Q_M\}$ and the reference image set $\mathcal{I}^R=\{\mathbf{I}^R_1, ..., \mathbf{I}^R_N \}$ from two different views represent two sets for the task, where the $M$ and $N$ denote the respective numbers of images. An embedding model is employed to obtain the feature representation for each set, which is denoted as $f_0:\mathcal{I} \rightarrow \mathbb{R}^{d_0}$. The output feature $ \mathbf{x}^k_i \in \mathbb{R}^{d_0}$ is a global vector with $d_0$ dimension and $k \in \{ Q, R \}$ denotes two different views.

In previous methods, the pre-trained backbone $b_0$ is employed as the initial $f_0$ and subsequently fine-tuned through supervised learning with the pair labels $S$. In our pipeline, we first freeze the parameters of the foundation model as $f_0$, and its output for each image is set as the initial features $\mathbf{x}^k_i$. We then train an adapter, serving as a bridge to map the initial features set $\mathcal{X}$ into a uniform feature space. The adapter is denoted by $f_\theta:\mathbb{R}^{d_0} \rightarrow \mathbb{R}^{d}$, where the $\theta$ denotes the learnable parameters of the adapter. The adapted feature $ \mathbf{z}^k_i \in \mathbb{R}^{d}$ is the output of the adapter.

In the test phase, the retrieval problem is defined as ranking the reference image set $\mathcal{I}^R$ by the feature similarities with the given query $\textbf{I}^Q \in \mathcal{I}^Q$. Since reference images contain GPS location, the top-ranked reference image's GPS is assigned to the query image for geo-localization.

\subsection{Pipeline Overview}
In the training of the geo-localization task, reference images from the satellite are typically curated and calibrated with location in advance. Subsequently, query images collected from cellphones or drones are labeled to corresponding reference images through annotation efforts. We eliminate the need for the expensive annotation process, relying solely on arbitrary unlabeled query images for training. However, due to the indeterminate relationship between the reference and query images, sampling the corresponding image pairs for normal supervised learning becomes unfeasible. 

As we have access to all the training reference data, we extract the initial features $\textbf{x}^R_i=f_0 ( \textbf{I}^R_i)$ from the entire reference image set $\mathcal{I}^R$. We then randomly select $M$ samples to extract the query initial features $\textbf{x}^Q_i = f_0 ( \textbf{I}^Q_i)$ from the training query data. The adapter maps the initial features for both reference and query images into a uniform feature space $\mathcal{Z}$, where the positive cross-view image pairs should exhibit high similarity. Since the positive pairs of unlabeled data are uncertain, we introduce an EM-based Pseudo-labeling (EMPL) module to effectively mine positive samples. We also propose the Adaptation Information Consistency (AIC) module to preserve the robust representation of the initial features. The EMPL and AIC modules accumulate the loss and iteratively update the adapter. The overview of the proposed self-supervised adaptation is illustrated in Fig.~\ref{fig:emb_frame}. 
\subsection{Foundation Model Feature Representation}
Recent methods of geo-localization focus on extracting the global feature via the deep learning models, which is easily achieved retrieval by calculating the feature similarity. Considering the robustness of the large-scale pre-trained model (coined Foundation Models~\cite{bommasani2021opportunities}), various foundation models are adopted as the backbone of the model for fine-tuning with supervised learning. Recently popular foundation models such as DINO, DINOv2, and CLIP are widely employed in the geo-localization task, due to their strong zero-shot capability. To aggregate the global vector feature from the local feature map of the model, we employ the Generalized Mean Pooling (GeM) on the value feature $\textbf{v} \in \mathbb{R}^{d_0}$ outputted on layer 31. Therefore, we define the global feature of the image as:
\begin{equation}
    \textbf{x} = \Big{(}\sum^{H\times W}_{i=1} \textbf{v}_i^p\Big{)}^{\frac{1}{p}},
\end{equation}
where $p=3$ represents GeM, $H$ and $W$ are the height and width of the local feature map, respectively.

Furthermore, we implement L2 normalization on the global features for enhanced training and retrieval stability. This process ensures that the features are distributed on the surface of the hypersphere, where the inner product between them aligns with the cosine similarity.

\subsection{EM-based Pseudo Labeling Module}
We aim to predict the matching probability between given queries and references, where the query should exhibit a high probability with the true reference of the same scene. Due to the metric inconsistency between cross-view initial features $\mathcal{X}^Q$ and $\mathcal{X}^R$, our objective is to learn an adapter that outputs uniform metric features $\mathcal{Z}^Q$ and $\mathcal{Z}^R$, where the similarity of these features accurately represents the matching probability. In this context, finding the optimal parameters of the adapter becomes crucial. Given the ground truth relationship label $S \in [0,1]^{M\times N}$ and observed data $\mathcal{X}$, the conventional approach maximizes the likelihood of the probability concerning adapter parameters $\theta$:

\begin{equation}\label{eq:sup_obj}
    p(\mathcal{X}, S|\theta),
\end{equation}
where the set $\mathcal{X}$ includes the $\mathcal{X}^Q$ and $\mathcal{X}^R$.

However, this quantity is intractable when $S$ is unobserved in our unlabeled procedure. Therefore, we need to estimate the likelihood with the same parameters $\theta$ without complete information on the data:
\begin{equation}\label{eq:unsup_obj}
    p(\mathcal{X}|\theta).
\end{equation}
Since labels are not observable for $\mathcal{X}$, we can treat this as a missing data problem and introduce the pseudo-labels $s$ as the latent variables. Therefore, we transform the Eq.~\ref{eq:unsup_obj} to an estimation of the following marginal likelihood:
\begin{equation}\label{eq:marginal}
    p(\mathcal{X}|\theta) = \int p(\mathcal{X}, s|\theta) ds.
\end{equation}

The pseudo-labels $s$ are utilized as an intermediate step toward the final matching prediction. Therefore, we propose to consider pseudo-labels as an implementation of the latent variables in Eq.~\ref{eq:unsup_obj}. However, training a model in this unsupervised manner presents a challenge as it involves simultaneously estimating optimal label predictions and model parameters. To address this complex learning problem, we decompose it by iteratively estimating the latent variables $s$ and the adapter parameters $\theta$. Therefore, the training of the adapter can be accomplished using the Expectation-Maximization (EM) algorithm~\cite{neal1998view}.

\begin{figure}
	\centering
	\includegraphics[scale=0.5]{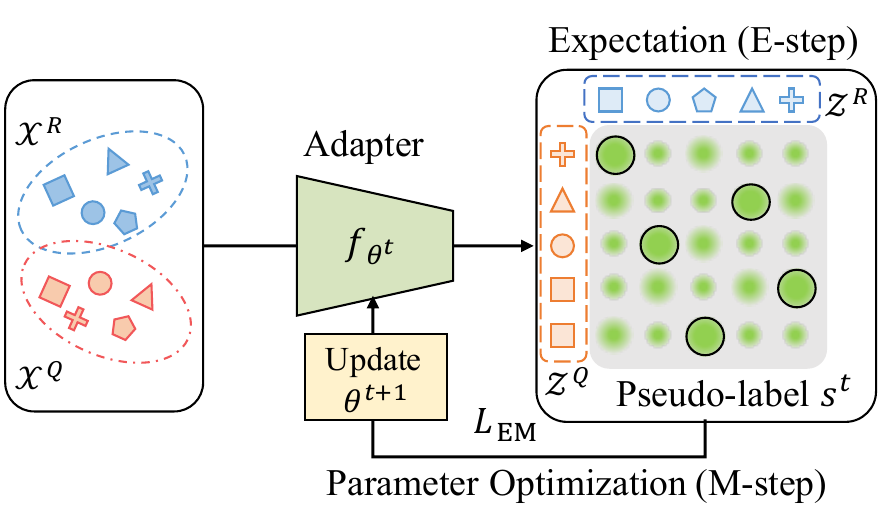}
	\caption{\textbf{Workflow of EMPL module.} The E-step is the pseudo-labeling of the positive pairs, while the M-step is updating the adapter with the supervision of the pseudo-labels.
	}
	\label{fig:EM}
\end{figure}

We introduce the EM-based Pseudo Labeling (EMPL) module, leveraging the EM algorithm to optimize both the adapter parameters $\theta$ and the distribution of hidden variables $s$, as depicted in Fig.~\ref{fig:EM}. The EMPL module involves two alternating steps to maximize the objective function in Eq.~\ref{eq:marginal} during training.

\textbf{E-step}: At the $t$-th iteration, the E-step estimates the posterior of the latent variable $s$ using the model $\theta^{(t)}$ updated from the last iteration. Following the cluster assumption~\cite{caron2020unsupervised} that similar data should have a high matching probability, the E-step performs inference on unlabeled initial features and calculates pseudo-labels $s^{(t)}$ based on the maximum predicted probability of the adapted features $Z^Q={\mathbf{z}^Q_1,...,\mathbf{z}^Q_M}$ and $Z^R={\mathbf{z}^R_1,...,\mathbf{z}^R_N}$. In practice, for this matching problem, the pseudo-label $s^{(t)}_{ij}$ for an adapted feature pair $\mathbf{z}^Q_i$ and $\mathbf{z}^R_j$ is assigned above a fixed threshold value, which is set at 0.1. This pseudo-labeling of assigning one-hot labels to the adapted features constitutes the E-step:

\begin{equation}\label{eq:pseudo-label}
s^t_{ij} = \begin{cases}
1, & \text{if } j = \arg\max_{j}  \langle \mathbf{z}^Q_i, \mathbf{z}^R_j \rangle\\
0, & \text{otherwise}.
    \end{cases}
\end{equation}

\textbf{M-step}: 
At the iteration $t$ of M-step, we optimize the optimal model parameters  $\theta^{t+1}$ by maximizing Eq.~\ref{eq:marginal}, where the latent variables $s^t$ are estimated from the E-step. This formulation can be solved using modern deep learning automatic differentiation. We utilize InfoNCE~\cite{oord2018representation} as the contrastive loss function during training:

\begin{equation}
    L_{\text{EM}}(Z^Q, Z^R, s^t) = - \frac{1}{M} \sum_{i=1}^M \log \frac{\sum_{j=1}^N s^t_{ij} \exp(\mathbf{z}^Q_i\cdot \mathbf{z}^R_j/\tau)}{\sum_{j=1}^N \exp (\mathbf{z}^Q_i \cdot \mathbf{z}^R_j /\tau)}.
\end{equation}

We optimize the adapter model by accumulating the loss function gradient and updating parameter $\theta^{t+1}$ with gradient descent.

\subsection{Adaptation Information Consistency Module}\label{sec:aic}
\begin{figure}
	\centering
	\includegraphics[scale=0.45]{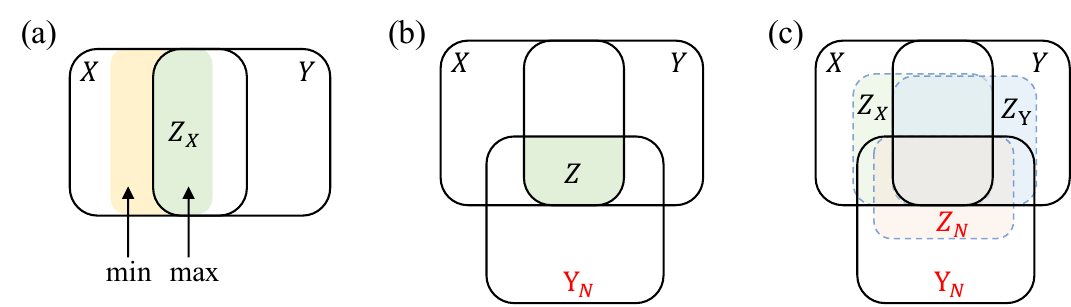}
	\caption{(a) Illustrates how contrastive learning compels $Z_X$ to extract mutual information and discard irrelevant information. (b) Demonstrates that if the matched target $Y$ is unknown, $Z$ may experience a decrease in mutual information and lack discriminativeness. (c) Emphasizes our goal to extract mutual information while preserving some redundancy information for enhanced robustness.
	}
	\label{fig:info-theory}
\end{figure}

Current works on cross-view geo-localization utilize supervised contrastive learning to train the model, where ground truth labels establish deterministic relationships for positive pairs of source samples $X$ and target samples $Y$. The objective of these learning paradigms is to maximize the mutual information of matched $X$ and $Y$ (depicted as the green part in Fig.~\ref{fig:info-theory}(a)), as well as minimize the task-agnostic information from original sources (depicted as the yellow part in Fig.~\ref{fig:info-theory}(a)).

However, our adaptation paradigm deals with the unlabeled CVGL task, where the targets $Y$ come from the uncertain pair predictions from the E-step. The potential false positive targets $Y_N$ may lead to a misguided optimization direction for the adapted feature. In this context, the uncertainty of positive pairs during training damages mutual information and reduces feature discrimination, as illustrated in Fig.~\ref{fig:info-theory}(b).

In our matching paradigm, the sources $X$ and targets $Y$ are the initial features of the query $X^Q$ and reference $X^R$, which are original robustness representations. The adapted features $Z^Q$ and $Z^R$ should preserve the redundancy information of the original features to improve the feature discriminativeness. We balance the redundancy information and scene-mutual information by maintaining feature consistency to improve the matching performance, as illustrated in Fig.~\ref{fig:info-theory}(c). Therefore, every adapted feature $Z$ should contain sufficient information that is consistent with the initial feature $X$.

\begin{figure}
	\centering
	\includegraphics[scale=0.45]{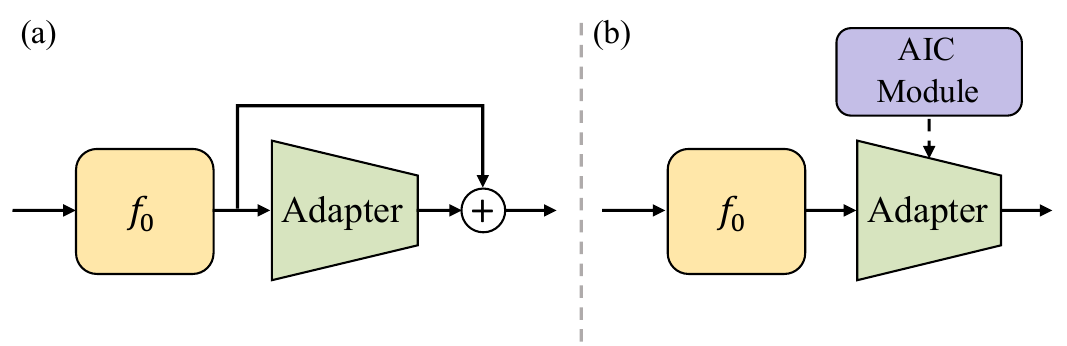}
	\caption{\textbf{Illustration of two architectures to maintain feature consistency.} (a) shows the residual style for the adapted features. (b) shows the AIC module regulates the adapter.}
	\label{fig:res-im}
\end{figure}

We introduce two architectures to ensure feature consistency. Firstly, we consider the Residual-style design~\cite{gao2023clip}, which straightforwardly aggregates the output of adapter and initial features $X$ as the final adapted feature, as illustrated in Fig.~\ref{fig:res-im}(a). Secondly, and more significantly, we propose the Adaptation Information Consistency (AIC) module to enforce feature consistency by regularizing the adapter during training, as depicted in Fig.~\ref{fig:res-im}(b).

\begin{figure}
	\centering
	\includegraphics[scale=0.5]{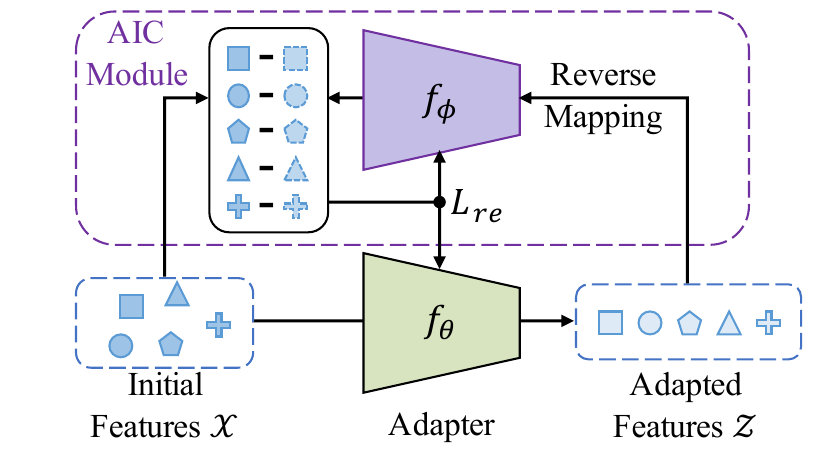}
	\caption{\textbf{Illustration of AIC Module.}
	}
	\label{fig:IM}
\end{figure}

The AIC module incorporates a reverse mapping network $f_\phi: \mathbb{R}^d \rightarrow  \mathbb{R}^{d_0}$  and a reconstruction loss $L_{\text{re}}$ to maintain the feature consistency, which is illustrated in Fig.~\ref{fig:IM}. The AIC module ensures that the adapted features encapsulate consistent information from the initial features. Inspired by the information bottleneck concept, we serve the adapter as an encoder that compresses information into the adapted features and subsequently reconstructs the initial feature through the reverter $\phi$. The reverter maps the adapted features $Z$ to the reconstructed features $\hat{X}$ that should be consistent with the initial feature $X$:
\begin{equation}
    \hat{X} = f_{\phi}(Z) = f_\phi(f_\theta(X)).
\end{equation}

The reconstruction loss function forces the reverted feature to be similar to the corresponding initial feature, reserving more information that is contained in the adapted feature. The loss function is the reconstruction loss that maintains the feature consistency:
\begin{equation}
    L_{\text{re}}(X, \hat{X}) = \sum_{\mathbf{x}_i \in X} || \mathbf{x}_i - \hat{\mathbf{x}}_i||^2_2.
\end{equation}
Note that the reverter is updated along with the adapter during the training phase.

\subsection{Overall Training and Inference Pipeline}

\SetKwComment{Comment}{/* }{ */}
\begin{algorithm}[hbt!]
\caption{Self-supervised adapter training pipeline}\label{alg: em}
\KwData{query initial feature set $X^{Q_0}$, reference initial feature set $X^R$,  training step $T$, query sampling number $M$}
\textbf{Initial}:adapter parameters $\theta^0$, reverter $\phi^0$\\
\textbf{Output}: learned adapter $\theta^T$\\
\For{$t=0, ..., T$}{
    $\textbf{Query Sampling}: $\\
    \quad $X^Q \gets RandomSample(X^{Q_0}, M)$\\
    $\textbf{E-step}: $\\
    \quad $Z^Q \gets f_{\theta^{t}}(X^Q) $\\
    \quad $Z^R \gets f_{\theta^{t}}(X^R) $\\
    \quad $s^t \gets f(Z^Q, Z^R) $ defined in Eq.~\ref{eq:pseudo-label}\\
    $\textbf{M-step}: $\\
    \quad $\hat{X}^Q \gets f_{\phi^{t}}(Z^Q)$\\
    \quad $\hat{X}^R \gets f_{\phi^{t}}(Z^R)$\\
    \quad $C = L_{\text{re}}(Z^Q, \hat{X}^Q) + L_{\text{re}}(Z^R, \hat{X}^R)$\\
    \quad $L = L_{\text{EM}}(Z^Q, Z^R, s) + L_{\text{EM}}(Z^R, Z^Q, s^\top)$\\
    \quad Update $\phi^{t+1}$ by descending $\partial C/ \partial \phi^t$ \\
    \quad Update $\theta^{t+1}$ by descending $\partial (L+C)/ \partial \theta^t$ \\
}
\end{algorithm}
Before the training phase for the adapter, all the initial features of images are extracted using the frozen model $f_0$. During training, we randomly sample $M$ query features $X^Q$ from the whole training query set $X^{Q_0}$. These sampled query features along with all the reference features $X^R$ are input into the adapter $f_\theta$ to obtain the adapted features $Z^Q$ and $Z^R$. The AIC module is coupled with the EMPL module to simultaneously update the parameters $\theta$ of the adapter and the parameters $\phi$ of the reverter. Specifically, the $L_{\text{EM}}$ and $L_{\text{re}}$ are accumulated for updating the adapter, while the $L_{\text{re}}$ of the AIC module also updates the reverter. The overall training pipeline is summarized in Alg.~\ref{alg: em}. 

During the inference phase, both the reference and query images are processed through the frozen model and input subsequently to the trained adapter, producing the global features of images. The similarities between the global features of the query and references are computed and ranked in descending order. As the reference images are GPS-tagged, the GPS location of the top-ranked reference is assigned as the location of the query image.

\section{Experiments}
\label{sec:experiment}

\subsection{Dataset and Evaluation Metric}
\textbf{University-1652 \& University-160k}:
The University-1652~\cite{zheng2020university} dataset is the pioneering drone-based geo-localization dataset, providing multi-view and multi-source images. The dataset comprises images from 1652 university buildings worldwide, where each building includes 1 satellite-view image and 54 drone-view images. The images of 701 buildings are utilized for training, while other 701 buildings are reserved for evaluation. To increase complexity, the reference images of the remaining 250 buildings are included in the reference test set as distractors. The geo-localization task involves matching drone-view images to corresponding satellite-view images and vice versa, with the evaluation metric being the average precision (AP). Zhang~\etal~\cite{zheng2023uavs} further introduced the University-160k dataset, an extension of the University dataset, by incorporating 160k satellite-view images as noise data to the original 951 satellite-view reference images during evaluation. This updated dataset serves to analyze the robustness and generalization capabilities of the feature representation.

\textbf{CVUSA \& CVACT}:
CVUSA is the first standard ground-to-satellite geo-localization dataset, while CVACT is another standard benchmark dataset. CVUSA and CVACT encompass 35,531 pairs of street-view panoramas and corresponding satellite-view images for training. The CVUSA provides 8,884 pairs for testing and CVACT has the same number of pairs in its validation set. CVACT additionally provides 92,802 pairs as its test set. The satellite images have a resolution of  $750 \times 750$, while the street-view counterparts have a resolution of  $224\times 1232$. Notably, the geographical north of street-view images aligns with the upper center of the satellite images. These one-to-one matching ground-to-satellite tasks are more challenging, as they encounter dramatic view discrepancies. 

\textbf{Evaluation Metrics}:
To evaluate the performance on the Geo-localization task, we use the Recall of Top-K in retrieval, denoted as R@$k$.  Additionally, we use Average Precision (AP) for the University-1652 dataset, which involves one-to-many and many-to-one matching. 
\begin{itemize}
    \item \textbf{Top-K Recall}: Top-K Recall (R@k) computes the number of queries where the ground truth label is among the top $k$ label prediction.
    \item \textbf{AP}: The Average Precision (AP) is the area under the Precision-Recall (PR) curve, which considers the one-to-many matching in the reference. We report the mean AP value of all queries in the experiments. 
\end{itemize}

\begin{table*}
	\caption{Comparsion between recent methods and our proposed method on the University-1652.}
	\label{tab:compare-u1652}
	\centering
	\setlength{\tabcolsep}{3mm}
		\begin{tabular}{c|ccc|cc|cc}
            \toprule
            \multirow{2}{*}{Method} & \multirow{2}{*}{Learning Type} & \multirow{2}{*}{Training Set} & \multirow{2}{*}{Training Parameters (M) } & \multicolumn{2}{c|}{Drone-to-satellite} & \multicolumn{2}{c}{Satellite-to-Drone} \\
            &&&&R@1 & AP &R@1 & AP \\ 
            \toprule
            ResNet-50~\cite{he2016deep} & \multirow{3}{*}{Generalization} & \multirow{3}{*}{-}&\multirow{3}{*}{-}  &5.95 & 8.13 & 21.53 & 11.24 \\
            DINOv2~\cite{oquab2024dinov} &&& & 31.25 &40.67 &66.48 & 37.91 \\
            
            AnyLoc~\cite{keetha2023anyloc} & & & &34.36 & 42.32 & 68.23 & 41.24 \\
            \midrule
            University-1652~\cite{zheng2020university}& \multirow{7}{*}{Supervised} & \multirow{7}{*}{Labeled Training Set}&26& 58.23&62.91 & 74.47 &	59.45\\
            RK-Net~\cite{lin2022joint}  & &&26 & 66.13   &70.23 &80.17	&65.76\\   
            LCM~\cite{ding2020practical}     & &&26 & 66.65   &70.82 &79.89 &65.38\\   
            DWDR~\cite{wang2022learning}    & &&26 & 69.77   &73.73 &81.46	&70.45\\   
            LPN~\cite{wang2021each}     & &&26 & 75.93   &79.14 &86.45 &74.79 \\   
            FSRA~\cite{dai2021transformer}    & &&30 & 82.25	&84.82	&87.87	&81.53 \\  
            Sample4Geo~\cite{Deuser_2023_ICCV}&&&36&92.65  &93.81  &95.14   & 91.39\\  
            \midrule
            Ours  &  \multirow{1}{*}{Self-supervised} &  \multirow{1}{*}{Unlabeled Training Set} & 2.25 & 70.29 &74.93 & 79.03 & 61.03 \\
			 \bottomrule
		\end{tabular}%
\end{table*}
\subsection{Implementation Details}
The adapter is a linear project matrix with size $d_0 \times d$ and the reverter is also a linear project matrix with size $d \times d_0$. The loss weights of $L_{\text{EM}}$ and $L_{\text{re}}$ are all set to 1.  We use the Adam optimizer with an initial learning rate of 0.001. We use a single NVIDIA RTX 4090 for all the experiments. 

To facilitate efficient adapter training, we save all the initial features through inference using the frozen models before the training of the adapter. In the experiments of University-1652 and University-160k datasets, the training iteration $T$ is set to 60. In each iteration, we input all initial features of satellite images and randomly select 700 samples from the 37,855 initial features of drone images. In the experiments of CVUSA and CVACT datasets, $T$ is set to 10. We randomly sample 1.5k initial features from ground and satellite images, respectively.

\subsection{Benchmarking Results}
\textbf{University-1652}:
We first evaluate the proposed method in the Drone-to-satellite geo-localization scenario. Since the proposed pipeline is the first method without labeled data, our evaluation involves a comprehensive comparison with both non-training (generalization) and supervised learning methods. The generalization models are pre-trained in a task-agnostic manner and evaluated in the downstream CVGL task. DINOv2 is the baseline foundation model in the experiment, and AnyLoc is the SOTA method that applies an unsupervised model with manually selected layers for the general geo-localization tasks. We also evaluate the performance of ResNet-50~\cite{he2016deep} pre-trained on ImageNet~\cite{deng2009imagenet} for comparison. The supervised methods are trained using the labeled training set of University-1652. Our proposed method is uniquely trained in the unlabeled training set, employing the proposed self-supervised paradigm. All these methods are evaluated in the test set of University-1652, which is summarized in Table~\ref{tab:compare-u1652}. 

First of all, the comparison between the foundation models and the pre-trained ResNet-50 underscores the effectiveness of the foundation models. In Drone-to-satellite retrieval, our method significantly enhances the R@1 accuracy of the foundation model DINOv2, escalating it from 31.25\% to an impressive 70.29\%, and increasing AP from 40.67\% to 74.93\%. Moreover, improvements are also observed in Satellite-to-Drone retrieval, where R@1 and AP are elevated from 66.48\% to 79.03\% and from 37.91\% to 61.03\%, respectively. Our proposed method demonstrates competitive performance when compared with recent supervised learning approaches. It achieves higher accuracy than several supervised methods in both Drone-to-satellite and Satellite-to-drone retrieval scenarios. Notably, the proposed method only trains 2.25 M parameters, a considerable reduction compared to the parameters trained by recent supervised methods.

\begin{table}
	\caption{Performance comparison between unsupervised methods and our proposed method on University-160k.}
	\label{tab:compare-160k}
	\centering
	\setlength{\tabcolsep}{1.3mm}
		\begin{tabular}{c|c|cccc}
			\toprule
			Method & Learning Type & R@1 & R@5 & R@10& AP \\ 
                \toprule
                DINOv2 (Baseline) & Generalization & 19.34& 40.89& 48.29& 24.58 \\
                AnyLoc & Generalization & 24.12 & 46.25 & 57.31& 31.53\\
                DINOv2 + Ours & Self-supervised & \textbf{39.47} & \textbf{81.01} & \textbf{87.90}& \textbf{48.96} \\
			 \bottomrule
		\end{tabular}%
\end{table}

\textbf{University-160k}:
We extend our evaluation to a more challenging Drone-to-satellite CVGL using the University-160k dataset. This dataset shares the same training set as the University-1652 dataset, but its test set introduces an additional 160k irrelevant satellite images as distractors in the reference. In Table~\ref{tab:compare-160k}, we compare our proposed methods with frozen foundation models. Despite the challenge posed by the increased distractors, our method outperforms the direct generalization of frozen foundation models. Notably, our method enhances the performance of DINOv2, improving Recall@1 from 19.34\% to 39.47\% and Average Precision (AP) from 24.58\% to 48.96\%.

\subsection{Results of Task-specific Pre-trained Model Adaptation}

The adapter refines features from the frozen foundation models on an unlabeled dataset, leveraging its self-supervised learning capability to mitigate the view gap. In situations where the models are pre-trained on a dataset for a specific CVGL task and then evaluated on a new dataset, our method can enhance the performance of these frozen task-specific pre-trained models by adapting the feature distribution to the new scenario. To empirically evaluate the effectiveness of our proposed method, we conduct experiments designed to assess its ability to improve the performance of task-specific pre-trained models when confronted with new, unlabeled data. This mirrors real-world scenarios, like deploying a pre-trained model in a new city. Typically, when faced with a decline in accuracy due to shifting contexts, the conventional approach involves fine-tuning the entire model using labeled data from the new city. In contrast, our proposed method offers a self-supervised adaptation solution, allowing for model adaptation even in the absence of labeled data for the new city.

Therefore, we conduct experiments to analyze the benefit of the proposed adaptation on the frozen task-specific model. For the Drone-to-satellite geo-localization scenario, we use the University-1652 dataset and randomly split the original test set into two parts: the unlabeled adaptation set and the final test set. We pre-train the task-specific models in the training set and freeze them as the frozen model $f_0$ in our pipeline. We then train the adapter using the unlabeled adaptation set. The whole model is evaluated in the final test set. There is no overlap between the adaptation set and the final test set, which guarantees that the scene information is not leaked to the adaptation set. For the Ground-to-satellite geo-localization scenario, we follow the setting of the previous works~\cite{chen2023sam}, which evaluate the transferring performance of a trained model in a new dataset. In this experiment, the frozen models are pre-trained on the training set of either CVUSA or CVACT. We assess their performance before and after the self-supervised adaptation on the test set of the other dataset (denoted as CVUSA $\rightarrow$ CVACT and CVACT $\rightarrow$ CVUSA). Our paradigm trains the adapter on the adaptation set without any labels, even in the presence of substantially different images across datasets. The experimental settings are presented in the Table~\ref{tab:set2}

\begin{table*}
	\caption{The dataset details of the adaptation on the task-specific pre-trained model.}
	\label{tab:set2}
	\centering
	\setlength{\tabcolsep}{2mm}
		\begin{tabular}{c|ccc|ccc}
            \toprule
            \multirow{2}{*}{Task} & \multicolumn{3}{c|}{Split Strategy}& \multicolumn{3}{c}{ID Number} \\ 
            & Pre-train& Adaptation& Test& Pre-train & Adaptation& Test\\ 
            \toprule
            University Test Adaptation & Training Set & Adaption Set& Test Set& 701 & 250 & 751 \\
            CVUSA$\rightarrow$ CVACT &  CVUSA Training  &  CVACT Training  &  CVACT Val  & 35531 & 35531 & 8884\\
            CVACT $\rightarrow$ CVUSA &  CVACT Training & CVUSA Training & CVUSA Test  & 35531 & 35531 & 8884\\
		\bottomrule
		\end{tabular}%
\end{table*}

\begin{table}
	\caption{Adaptation of task-specific pre-trained model on the Drone-to-satellite dataset.}
	\label{tab:u1652-adapt}
	\centering
	\setlength{\tabcolsep}{2mm}
		\begin{tabular}{c|cccc}
			\toprule
			Method & Adaptation &  R@1 & R@5 & AP \\ 
            \toprule
            \textbf{University-1652 Split Test} \\
            FSRA       & $\times$     &  81.55&	93.50  & 84.25 \\
            FSRA       & $\checkmark$ &  83.83& 94.29  & 86.19\\
            Sample4Geo & $\times$     &  94.07& 95.28  & 95.78\\
            Sample4Geo & $\checkmark$ &  \textbf{95.84}& \textbf{98.61}  & \textbf{96.49}\\
		\bottomrule
		\end{tabular}%
\end{table}
\textbf{University-1652}:
The original test set of University-1652 is randomly split into an adaptation set and a final test set. Baseline models are pre-trained in the original training set, each following their respective strategies. We executed adaptation by training our adapter using the unlabeled adaptation set and subsequently evaluated the models in the final test set. To evaluate the impact of our proposed adaptation, we conducted a comparative analysis to compare performance before and after the adaptation process. We selected the FSRA and Sample4Geo models for benchmarking, both renowned methods in cross-view geo-localization. Detailed experimental results are outlined in Table~\ref{tab:u1652-adapt}.

The experimental results suggest that our proposed adaptation proves effective in enhancing performance when the baseline models are pre-trained for this task. Given the minimal discrepancy between the training set and the test set, the impact of our adaptation is observed to be slight in performance improvement.

\begin{table}
	\caption{Adaptation of task-specific pre-trained model on the Ground-to-satellite datasets.}
	\label{tab:g2s-adapt}
	\centering
	\setlength{\tabcolsep}{2mm}
		\begin{tabular}{c|cccc}
            \toprule
            Method & Adaptation &  R@1 & R@5 & R@1\% \\ 
            \toprule
            \textbf{CVUSA} $\rightarrow$ \textbf{CVACT}\\
            GeoDTR~\cite{zhang2023cross}     &$\times$     & 54.04&75.62&93.80\\
            GeoDTR &$\checkmark$     & 57.32&78.94& 95.10\\
            Sample4Geo &$\times$     & 56.62&77.79&94.69\\
            Sample4Geo &$\checkmark$ & \textbf{60.94}&\textbf{80.48}&\textbf{95.51}\\
		\toprule	
            \textbf{CVACT} $\rightarrow$ \textbf{CVUSA}\\
            GeoDTR     & $\times$      &44.07&64.66&90.09\\
            GeoDTR & $\checkmark$      &44.97&65.01&91.27 \\
            Sample4Geo & $\times$      &44.95&64.36&90.65 \\
            Sample4Geo & $\checkmark$  &\textbf{45.49}&\textbf{66.19}&\textbf{92.12}\\
			\bottomrule
		\end{tabular}%
\end{table}

\textbf{CVUSA \& CVACT}:
To evaluate the efficiency of the proposed adapter on the ground-to-satellite task across different cities, we conduct experiments across the CVUSA and CVACT. This experimental setup focuses on evaluating the transferring ability of models. Owing to the self-supervised learning on unlabeled data, our proposed method exhibits flexibility to adapt the task-specific pre-trained model through the unlabeled training set from another dataset, functioning as an adaptation set. The experimental results are shown in Table~\ref{tab:g2s-adapt}. 

Comparative results reveal that, given access to unlabeled data, our method can markedly enhance the performance of the pre-trained models when confronted with a novel city scenario. Given the existence of the view discrepancies among query ground images between different datasets, the feature distributions exhibit variations. While the pre-trained methods may experience performance degradation in such scenarios, our proposed paradigm excels in adapting the frozen models to unify the feature representation between query and reference features within new datasets.

\subsection{Ablation Study}\label{sec:ablation}

\textbf{Effectiveness of the Proposed Modules}:
To evaluate our design of the self-supervised method, we conduct an ablation study in the University-1652.
We analyze the impacts of the proposed modules and explore their effectiveness on different configurations to better understand their contributions to this task. Additionally, we compare different designs to preserve the robustness of the frozen model, including the proposed AIC module and the residual-style design, which is presented in Table~\ref{tab:ablation}.

Initially, we directly generalize the frozen foundation model without adaptation as the baseline, which is listed in the first column in the table. We train the adapter using labeled data as supervision to analyze the effect of the supervised adaptation. Although the supervised adaptation converges quickly during training, it makes the adapter overfit the training data. We then apply the proposed EMPL module and train the adapter with unlabeled data. This design gains a superior improvement in retrieval accuracy than the supervised counterpart. This comparison highlights the importance of the EMPL, which can improve performance even without label data.

We also compared the impact of the proposed AIC module with the Residual-style design. Since the residual connection necessitates identical dimensions for both input and output features, for a fair comparison, the adapter trained with the AIC module maintains the same output feature dimension as the input. The experimental comparison is presented in the last two columns of Table~\ref{tab:ablation}. Remarkably, the results indicate that the residual connection may impose limitations on the adapter's performance, resulting in a decline in accuracy. In contrast, the proposed AIC Module further provides an enhancement of accuracy by incorporating the reconstruction loss, even when maintaining the same dimension as the input features.

\begin{table}[t!]
    \centering
	\caption{Ablation study of the proposed modules on University-1652 dataset.}
	\label{tab:ablation}
	\setlength{\tabcolsep}{1.7mm}
		\begin{tabular}{c|c|cccc}
			\toprule
			Module & Baseline & \multicolumn{4}{c}{Different Configurations} \\ 
                \midrule
                Supervised      &  &\checkmark & &  &  \\
			    EMPL   &  & &\checkmark &\checkmark &\checkmark \\
			    Residual-style  &  &  &  &\checkmark & \\
                AIC       &  &  &  &  &\checkmark \\
                \midrule
			    R@1   &31.60 &59.61&63.93&55.44&\textbf{69.45}\\
			    AP    &36.60 &70.89&73.79&60.85&\textbf{74.07}\\ 
                \bottomrule
		\end{tabular}%
\end{table}

\textbf{Choices of Adapted Feature Length}:
We further analyze the effect of the feature embedding length $d$ of the adapter. The AIC module employs the loss function to force the adapted feature to preserve the quality of the initial features, which does not have the equal constraint of the dimension $d_0$ of initial features and the dimension $d$ of adapted features. Therefore, it can have various output sizes of the adapted features. We analyze the different performances of the setting of the dimension of the adapted features. We fix the input size $d_0=1536$ of the initial feature from the foundation model and evaluate different output dimensions of the adapter, which is presented in Table~\ref{tab:dim}. The accuracy improves as the output size dimension of the adapter increases. Considering both retrieval effectiveness and storage cost, the output size of $d = 2048$ is a trade-off selection for our proposed method.

\begin{table}
	\caption{Ablation study on the embedding length of the adapter.}
	\label{tab:dim}
        \centering
	\setlength{\tabcolsep}{1mm}
		\begin{tabular}{c|c|cccccc}
            \toprule
            Dimension & 1536 (base) & 512  & 1024 & 1536 & 2048 & 2560 & 3072 \\ 
            \midrule
            R@1 & 31.25&65.62 & 67.38& 69.45& 70.29& 70.24& \textbf{70.73} \\
            AP &40.67 &70.56&72.25&74.07&74.93&74.83&\textbf{75.24} \\
            \bottomrule
		\end{tabular}%
\end{table}

\textbf{Effectiveness of the Adapter}:
We offer an insight into the proposed pipeline's impact on the frozen model. Initially, we employ the foundation models to direct generalization within the single-view geo-localization task. Specifically, we opt for a Drone-to-drone single-view setup within the University-1652 dataset. We select a single drone image as the reference of each scene, while the remaining images serve as queries. Notably, this setup eliminates the view discrepancy between the query and reference images, showcasing the upper limits of the foundation models' feature representation capabilities. Subsequently, we proceed to evaluate and compare the performance of the foundation models in both direct generalization and the proposed adaptation within a cross-view setup. This experimental design involves assessing the models' capabilities in handling view discrepancies between query and reference images, providing valuable insights into the efficacy of the proposed adaptation approach. The results are represented in Table~\ref{tab:improve-ablation}.

From the Drone-to-drone scenario to the Drone-to-satellite scenario, the baseline DINOv2 suffers from degradation of 71.99\% to 31.24\% in R@1 and 81.36\% to 40.67\% in AP. When employing our adaptation in the cross-view task, the performance of DINOv2 rebounds remarkably to 70.29\% in R@1 and 74.93\% in AP, approaching the levels achieved in the single-view task. This trend is also observed in the similar results for CLIP. This experiment underscores the pivotal role of the proposed adaptation in mitigating the impact of view discrepancies, effectively restoring the robust feature representation inherent in foundation models.

\begin{table}
	\caption{Generalization ablation of the proposed method.}
	\label{tab:improve-ablation}
	\centering
	\setlength{\tabcolsep}{2mm}
		\begin{tabular}{c|c|cccc}
			\toprule
			Model &Task& R@1 &R@5& R@10& AP \\
			\toprule
            CLIP	     &\multirow{2}{*}{Drone-to-drone}	 &40.12  &66.04 &75.55 &46.12 \\
            Dinov2	     &      &\textbf{71.99}	&\textbf{93.30} &\textbf{96.34} &\textbf{81.36} \\
            \midrule
            CLIP	     & \multirow{4}{*}{Drone-to-satellite} &12.30 &25.78&33.91 &16.06\\
            Dinov2	     & &31.24 &50.48 &58.61 &40.67 \\
            CLIP+Ours	 & &13.58  &28.46 &37.19  &17.60 \\
            Dinov2+Ours	 & &\textbf{70.73}&	\textbf{91.83}	&\textbf{95.69}&	\textbf{75.24} \\
			\bottomrule
		\end{tabular}%
\end{table}

\begin{figure}
    \centering
    \includegraphics[width=0.45\textwidth]{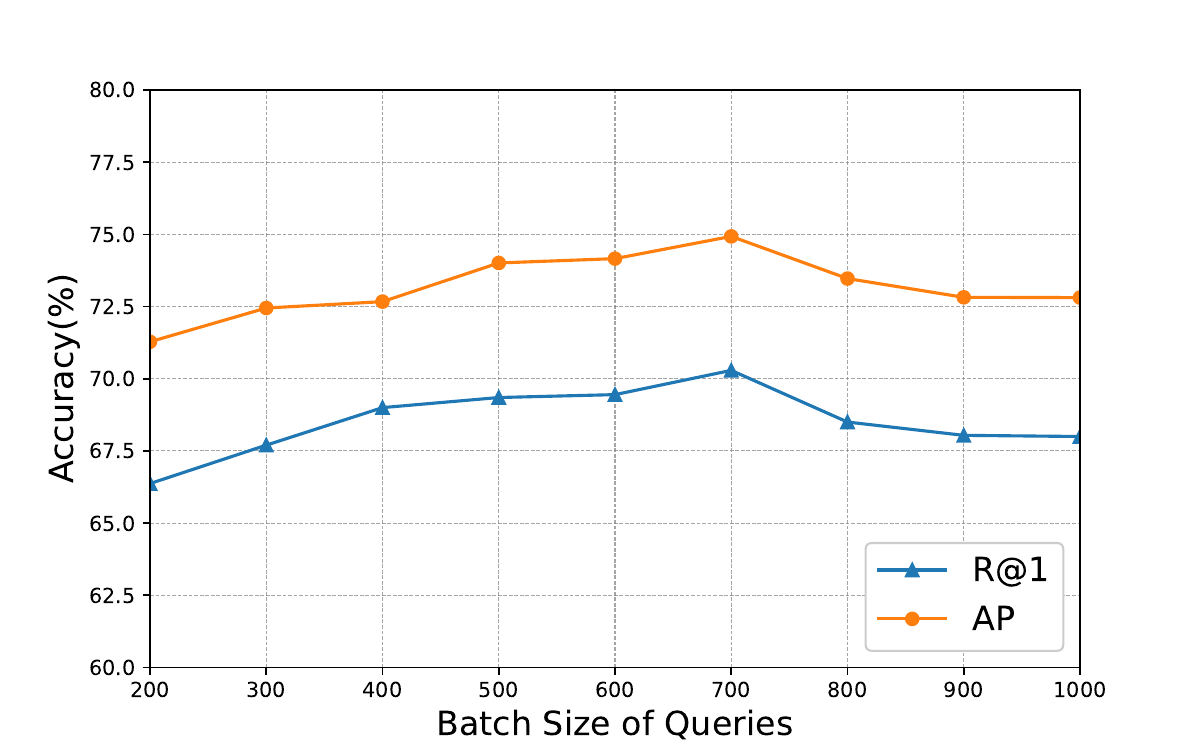}
    \caption{The performance on University-1652 under different batch sizes of queries during training.}
    \label{fig:batch_size}
\end{figure}

\textbf{Effects of the Batch Size}:
The training pipeline for the adapter utilizes the contrastive learning loss function $L_{\text{EM}}$ to optimize the adapter. As contrastive learning involves negative samples from a minibatch~\cite{chen2022we}, the batch size significantly impacts the performance. Therefore, we focus on analyzing the impact of the batch size of query samples in the training pipeline. As we can access all reference satellite data during adaptation, we set all 701 satellite training samples in one iteration and investigated the batch size of queries. The results depicted in Fig.~\ref{fig:batch_size} indicate that the performance is stable among different sizes and reaches its peak when the number of queries is equal to that of the references. This study suggests that our proposed pipeline exhibits less sensitivity to variations in the batch size of queries.

\begin{figure*}
    \centering
    \includegraphics[width=0.9\textwidth]{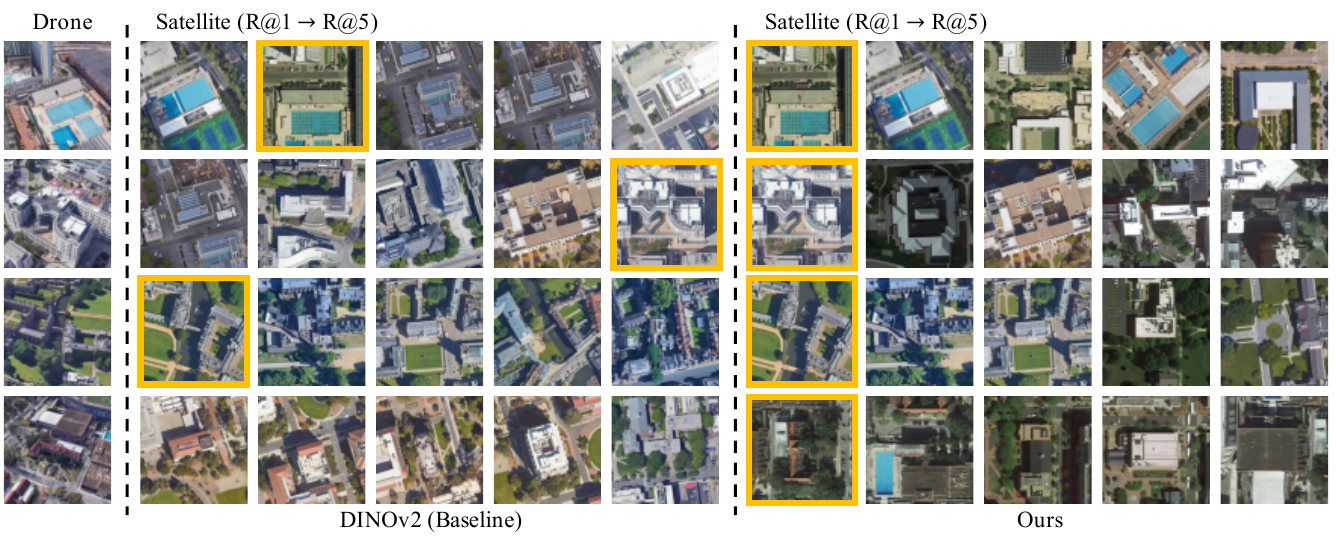}
    \caption{\textbf{Results of the Drone-to-satellite retrieval result on the University-1652 Dataset.} The query drone images are listed on the left, and the yellow boxes indicate the true matched images.}
    \label{fig:well_vis_d2s}
\end{figure*}

\begin{figure*}
    \centering
    \includegraphics[width=0.9\textwidth]{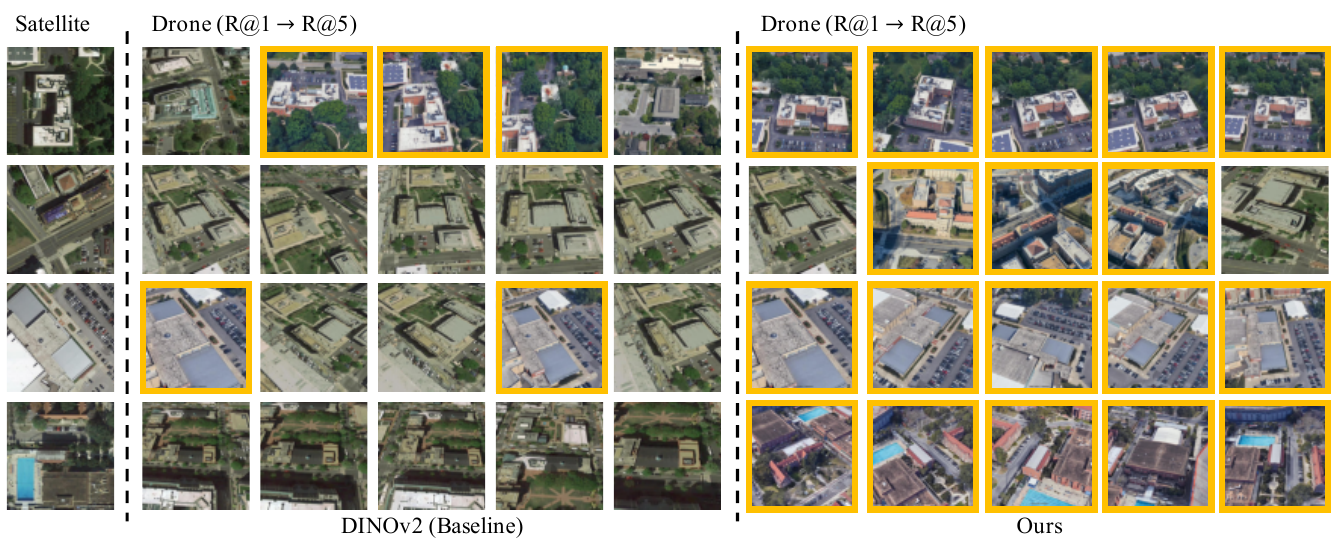}
    \caption{\textbf{Results of the Satellite-to-drone retrieval result on the University-1652 Dataset.} The query satellite images are listed on the left, and the yellow boxes indicate the true matched images.}
    \label{fig:well_vis_s2d}
\end{figure*}

\subsection{Visualization}
\textbf{Visualization of Retrieval Results}:
We present the visualization of retrieval and match results on University-1652  and compare them with the foundation model DINOv2. Given the drone image queries, as shown in Fig.~\ref{fig:well_vis_d2s}, DINOv2 potentially retrieves the textual similar target. After the adaptation, the model can precisely retrieve the true similar context targets. One reason is that the FM considers the textual information, however, suffers from the view gap between the queries and references. Our proposed model can lead the initial feature of the FM to overcome the gap and achieve better performance. The result of the Satellite-to-Drone scenario is shown in Fig.~\ref{fig:well_vis_s2d}. The result of the baseline reveals that query satellite images mostly correspond to an incorrect scene, as depicted in row 2, where the drone images of this misidentified scene share a similar color and style with the satellite queries. The ground truth targets may exhibit a different style but contain the same contextual elements within the scene. Consequently, the foundation model tends to misalign targets based on style similarity. We further analyze this situation in Fig.~\ref{fig:well_vis_s2d}. Compared with the foundation model, the proposed adapter can match the true positive targets even if the query and reference images have different colors and styles, demonstrating an ability to alleviate both view and style discrepancies.

\begin{figure}
    \centering
    \includegraphics[width=0.5\textwidth]{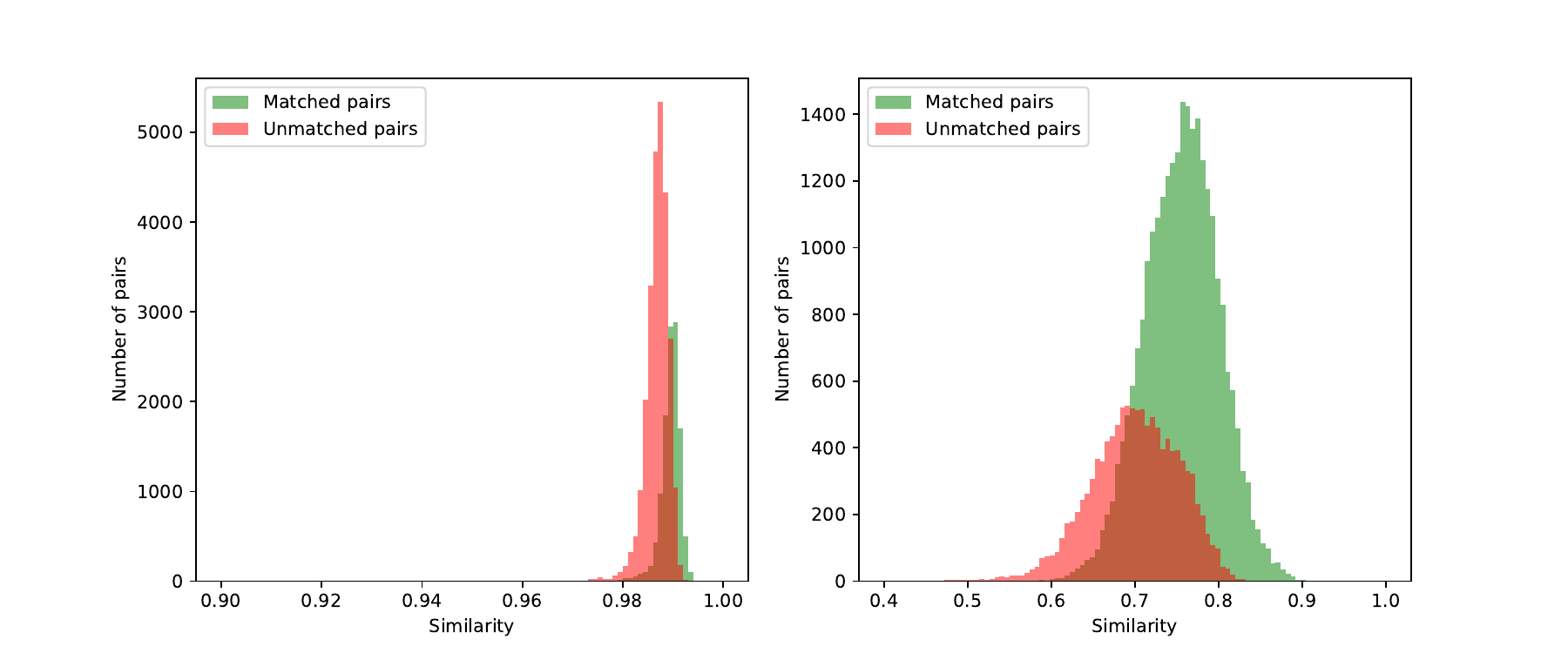}
    \caption{\textbf{Illustration of pairwise similarity distributions.} We compare the pairwise similarity distributions of \textbf{matched pairs} vs. \textbf{unmatched pairs} for both initial features (\textbf{left}) and adapted features (\textbf{right}) on the test set.}
    \label{fig:tp-fp}
\end{figure}

\textbf{Visualization of Similarity Histograms}:
To visually demonstrate the enhancements brought by our proposed method, we present pairwise similarity histograms of the University-1652. Specifically, the matched pairs constitute the true positives identified in top-1 predictions, while the unmatched pairs represent the false positive pairs. As depicted in Fig.~\ref{fig:tp-fp}, we show the histograms of both initial features (left) and adapted features through our proposed adaptation (right). Upon inspecting the left figure, it is apparent that the distributions of initial features are relatively narrow, primarily centered around a similarity range of 0.98 to 1.0. This characteristic makes it challenging to effectively differentiate between true and false image pairs. In contrast, our proposed method adapts the features, significantly improving the discernibility between positive and negative pairs. The adapted features exhibit a more extensive range in the histogram on the right, with positive pairs consistently displaying higher similarity values than the negative pairs. While some overlap persists between the matched and unmatched pairs, our method proves effective in enhancing performance, primarily due to a lower proportion of negative pairs and a larger number of matched pairs compared to the initial features. This experiment demonstrates the efficacy of our proposed adaptation in augmenting the model's ability to discriminate between positive and negative pairs.

\begin{figure}
    \centering
    \includegraphics[width=0.5\textwidth]{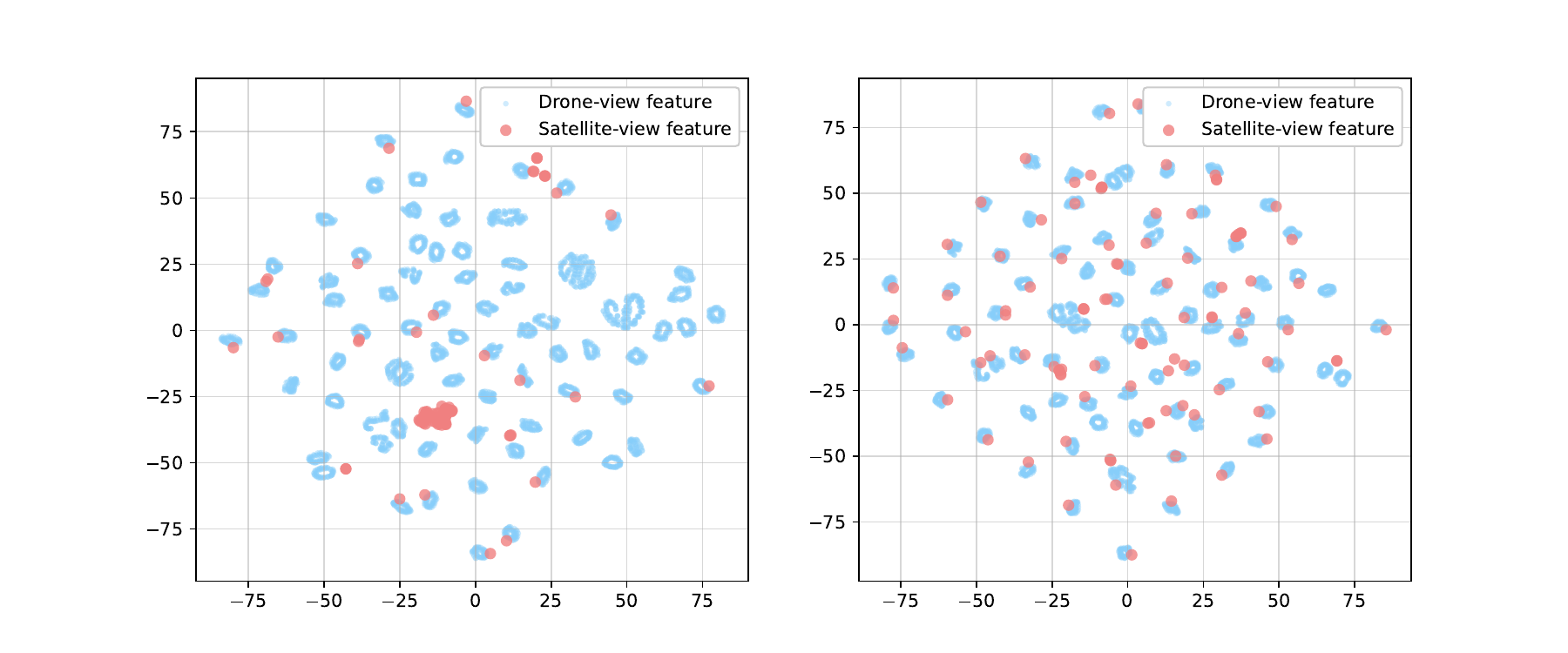}
    \centering
    \caption{\textbf{Illustration of t-SNE results.} We visualize the distributions of drone-view and satellite-view features from the foundation model (\textbf{left}) and the proposed method (\textbf{right}) via t-SNE. }
    \label{fig:tsne}
\end{figure}

\textbf{Visualization of Feature Distributions}:
We further present a visualization of the feature distribution for both drone and satellite images in the University-1652, comparing the initial features obtained from the foundation model with those enhanced by our proposed method. In Fig.~\ref{fig:tsne}, we depict the two-dimensional embeddings of satellite and drone image features, utilizing t-SNE for visualization. Specifically, we focus on 100 randomly selected scenes within the test set. The red circles denote features of satellite images, each corresponding to a distinct scene, while blue dots represent drone image features, with each scene encompassing 54 drone image features. Overlapping features signify their association with the same scene.

Examining the distribution of initial features on the left, we observe that drone images of the same scene are close to each other, indicating the foundation model's robust feature representation in the single-view context. However, the satellite features exhibit less overlap with their corresponding drone features, leading to a degradation in accuracy. Additionally, satellite features from different scenes are closer to each other, representing a cluster in the figure. This clustering effect results in satellite features wrongly matching the same drone target that is close to this cluster, leading to the situation shown in Fig.~\ref{fig:well_vis_s2d}. On the right, we present the distribution of adapted features through our proposed method. Notably, all corresponding satellite and drone images are clustered together, underscoring the efficiency of the proposed adaptation in bridging the view gap. Drone features from the same scene exhibit proximity, preserving the robustness of the foundation model. This visualization underscores the adaptability of our method in aligning features across different perspectives, addressing the view discrepancy challenge present in the initial foundation model.

\begin{figure}
    \centering
    \includegraphics[width=0.4\textwidth]{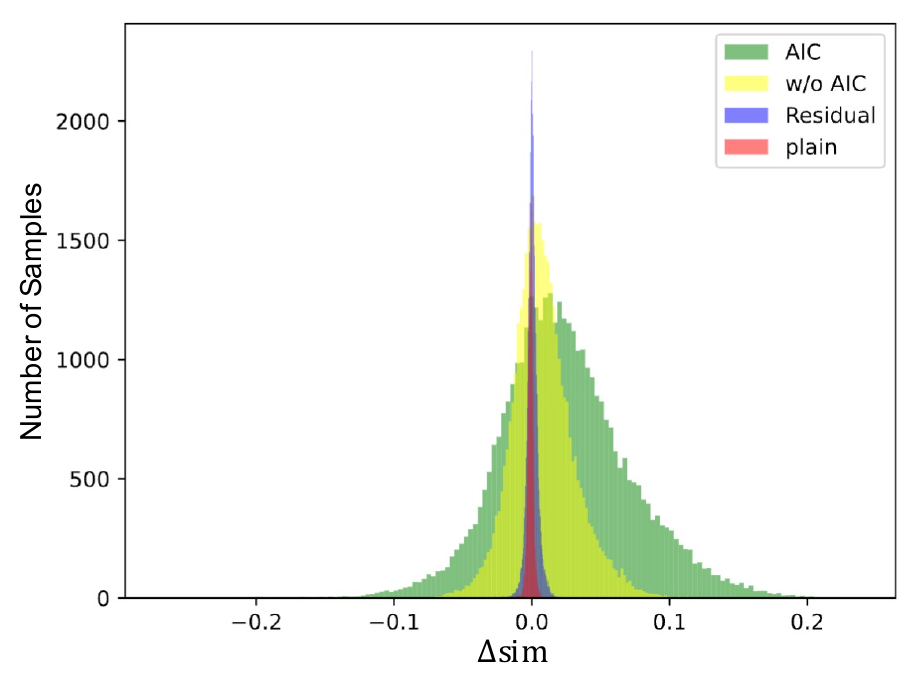}
    \centering
    \caption{\textbf{Histogram for the similarity differences on University-1652.} We show the differences $\Delta \text{sim}$ between true similarities $\text{sim}_T$ and hard negative similarities $\text{sim}_N$ on test set under various architectures.}
    \label{fig:sim}
\end{figure}

\textbf{Analysis of the AIC Module}:
In Fig.~\ref{fig:sim}, we analyze the impact of the AIC Module by visualizing the similarity difference $\Delta\text{sim} = \text{sim}_T - \text{sim}_N$. The $\text{sim}_T$ represents the similarity of ground true pairs, and $\text{sim}_N$ represents the similarity of hard negative pairs. A larger value of the difference indicates more discriminative features. Both plain initial features and Residual-style features exhibit small $\Delta\text{sim}$ values. Because the Residual-style adapter only learns an offset of the features, thereby inheriting poor discriminativeness. Although the adapter without AIC enhances matching accuracy, it results in low discriminative features. This supports the hypothesis illustrated in Fig.~\ref{fig:info-theory}(b), indicating oversimilarity between adapted features. When integrating the AIC module, the features exhibit increased discriminativeness, leading to an improved ratio of positive samples and a noticeable rightward shift in the histogram. These results support our motivation in Section~\ref{sec:aic}, asserting that redundant information contributes to discriminative features. Additionally, it is noted that there are still negative samples with larger similarities than true positives due to potential negative targets being matched during training on unlabeled data.

\section{Discussion}
\label{sec:discussion}

Our comprehensive experiments reveal the significant potential of self-supervised adaptation techniques for frozen models in CVGL, where both the strong frozen model and the self-supervised adapter emerge as pivotal components contributing to remarkable performance. These findings raise important questions about the generalizability of frozen models across diverse CVGL datasets and the consistent enhancement of adapters across various frozen pre-trained model architectures.

\subsection{Generalizability of Pre-trained Models Across Datasets}
\label{sec:dis1}

Upon scrutinizing the performance of pre-trained models across different CVGL datasets, we observe varying levels of success. While the task-specific pre-trained model necessitates training in the downstream task, limiting its generalizability, a more effective approach is to leverage the foundation model for all CVGL tasks. Although the foundation model exhibits considerable performance in the Drone-to-satellite scenario, transferring its capabilities to the Ground-to-Satellite scenario proves significantly challenging. Notably, the foundation model struggles to extract meaningful feature representations for cross-view retrieval, resulting in dismal results when generalized to datasets such as CVUSA and CVACT, where Recall@1 (lower than 0.1\%) and Average Precision (lower than 1\%)  plummet to negligible levels.

This suboptimal performance can be attributed to two primary factors. Firstly, panoramic ground-view photos exhibit a distorted imaging appearance, deviating from the feature distribution that the foundation model is adept at handling. Secondly, the view discrepancy in the Ground-to-Satellite scenario is considerably more challenging than that in the Drone-to-satellite task. This discrepancy makes it challenging for the model to establish correspondences between queries and references. Consequently, the foundation model's poor performance exacerbates the challenges in adjusting the adapter through our proposed training pipeline. This highlights the need for tailored solutions and adaptations to address the intricacies of the Ground-to-Satellite geo-localization task.

\subsection{Improvement of Adapters Across Different Frozen Models}

Our results presented in Table~\ref{tab:improve-ablation} reveal a noteworthy outcome: the proposed adapter successfully improves the accuracy of the foundation model DINOv2 to nearly its original levels. However, the adaptation performance for models such as CLIP (31.24\% in R@1 and 40.67\% in AP) still falls short of its maximum potential (40.12\% in R@1 and 46.12\% in AP) achieved in a single-view task. This discrepancy prompts a two-fold consideration. Firstly, the reliance on pseudo-labeling, contingent upon the frozen foundation model's accuracy, raises the hypothesis that the upper limitation of the proposed adaptation is inherently linked to the performance of the frozen model. Secondly, it hints at potential alternative methods for restoring the feature representation of foundation models in cross-view scenarios, warranting exploration in future research endeavors.

\subsection{Limitations and Future Works}

While our proposed method has shown promising performance in CVGL, its applicability of adapting the foundation model to Ground-to-Satellite geo-localization remains limited. Future efforts will focus on mitigating the view discrepancy between panoramic images and satellite images, potentially through pre-transformation techniques applied to images before feeding them into the foundational model. Moreover, extending our consideration beyond the inter-relationships between query and reference images to include intra-relationships within each distribution holds promise for further refining model understanding. Specifically, exploring the intra-relationship within the reference distribution during the training process could optimize the adapter's performance, because of the clear location of the reference images. Furthermore, leveraging the drone sensor's capacity to capture multiple images of a single scene opens avenues for generating novel perspectives through multi-view stereo techniques. Integrating multi-view data into the self-supervised learning pipeline can potentially eliminate view discrepancies or offer valuable sample augmentation for training purposes. With the University-1652 dataset providing multiple drone images per scene, future research will explore the integration of multi-view data to enhance the model's adaptability and robustness.

\section{Conclusion}
\label{sec:conclusion}
In this paper, we present a novel self-supervised adaption method aimed at addressing the challenge of cross-view geo-localization without ground truth. Our approach leverages the powerful feature extraction capabilities of frozen models while introducing a lightweight, self-supervised adapter module to bridge the inherent visual gap between diverse viewpoints. Key to our success is the efficient utilization of unlabeled data through an EM-based Pseudo-Labeling module that estimates associations between query and reference features, and an Adaptation Information Consistency module that preserves the rich information captured by the frozen models. Extensive experiments on both Drone-to-satellite and Ground-to-satellite datasets demonstrate that our method significantly enhances the performance of pre-trained models and achieves competitive accuracy compared to recent supervised methods. Notably, our method requires fewer training parameters, resulting in a substantial reduction of computational and time costs. Furthermore, the self-supervised nature of our adapter module allows for easy integration with task-specific pre-trained models, offering a plug-and-play solution for real-world applications. This flexibility is particularly valuable when deploying models in new cities, where labeled data might be scarce, making our method a highly practical and adaptable tool for various cross-view geo-localization tasks.

\section*{Acknowledgment}
The authors extend their sincere gratitude for their efforts in reviewing this paper. We also thank our colleagues for their valuable suggestions. Furthermore, the numerical calculations presented in this article were performed on the supercomputing system at the Supercomputing Center of Wuhan University.

\bibliographystyle{IEEEtran}
\small
\bibliography{ref} 

\begin{thebibliography}{10}
\providecommand{\url}[1]{#1}
\csname url@samestyle\endcsname
\providecommand{\newblock}{\relax}
\providecommand{\bibinfo}[2]{#2}
\providecommand{\BIBentrySTDinterwordspacing}{\spaceskip=0pt\relax}
\providecommand{\BIBentryALTinterwordstretchfactor}{4}
\providecommand{\BIBentryALTinterwordspacing}{\spaceskip=\fontdimen2\font plus
\BIBentryALTinterwordstretchfactor\fontdimen3\font minus \fontdimen4\font\relax}
\providecommand{\BIBforeignlanguage}[2]{{%
\expandafter\ifx\csname l@#1\endcsname\relax
\typeout{** WARNING: IEEEtran.bst: No hyphenation pattern has been}%
\typeout{** loaded for the language `#1'. Using the pattern for}%
\typeout{** the default language instead.}%
\else
\language=\csname l@#1\endcsname
\fi
#2}}
\providecommand{\BIBdecl}{\relax}
\BIBdecl

\bibitem{ren2023hashing}
P.~Ren, Y.~Tao, J.~Han, and P.~Li, ``Hashing for geo-localization,'' \emph{IEEE Transactions on Geoscience and Remote Sensing}, vol.~61, pp. 1--13, 2023.

\bibitem{xue2022terrain}
X.~Wan, Y.~Shao, S.~Zhang, and S.~Li, ``Terrain aided planetary uav localization based on geo-referencing,'' \emph{IEEE Transactions on Geoscience and Remote Sensing}, vol.~60, pp. 1--18, 2022.

\bibitem{lu2023okay}
X.~Lu, S.~Luo, and Y.~Zhu, ``It’s okay to be wrong: Cross-view geo-localization with step-adaptive iterative refinement,'' \emph{IEEE Transactions on Geoscience and Remote Sensing}, vol.~60, pp. 1--13, 2022.

\bibitem{sun2023cross}
Y.~Sun, Y.~Ye, J.~Kang, R.~Fernandez-Beltran, S.~Feng, X.~Li, C.~Luo, P.~Zhang, and A.~Plaza, ``Cross-view object geo-localization in a local region with satellite imagery,'' \emph{IEEE Transactions on Geoscience and Remote Sensing}, vol.~61, pp. 1--16, 2023.

\bibitem{zheng2020university}
Z.~Zheng, Y.~Wei, and Y.~Yang, ``University-1652: A multi-view multi-source benchmark for drone-based geo-localization,'' in \emph{Proceedings of the 28th ACM International Conference on Multimedia}, 2020, pp. 1395--1403.

\bibitem{lin2022joint}
J.~Lin, Z.~Zheng, Z.~Zhong, Z.~Luo, S.~Li, Y.~Yang, and N.~Sebe, ``Joint representation learning and keypoint detection for cross-view geo-localization,'' \emph{IEEE Transactions on Image Processing}, vol.~31, pp. 3780--3792, 2022.

\bibitem{wang2021each}
T.~Wang, Z.~Zheng, C.~Yan, J.~Zhang, Y.~Sun, B.~Zheng, and Y.~Yang, ``Each part matters: Local patterns facilitate cross-view geo-localization,'' \emph{IEEE Transactions on Circuits and Systems for Video Technology}, vol.~32, no.~2, pp. 867--879, 2021.

\bibitem{dai2021transformer}
M.~Dai, J.~Hu, J.~Zhuang, and E.~Zheng, ``A transformer-based feature segmentation and region alignment method for uav-view geo-localization,'' \emph{IEEE Transactions on Circuits and Systems for Video Technology}, vol.~32, no.~7, pp. 4376--4389, 2021.

\bibitem{zhao2024transfg}
H.~Zhao, K.~Ren, T.~Yue, C.~Zhang, and S.~Yuan, ``Transfg: A cross-view geo-localization of satellite and uavs imagery pipeline using transformer-based feature aggregation and gradient guidance,'' \emph{IEEE Transactions on Geoscience and Remote Sensing}, vol.~62, pp. 1--12, 2024.

\bibitem{workman2015wide}
S.~Workman, R.~Souvenir, and N.~Jacobs, ``Wide-area image geolocalization with aerial reference imagery,'' in \emph{Proceedings of the IEEE International Conference on Computer Vision}, 2015, pp. 3961--3969.

\bibitem{liu2019lending}
L.~Liu and H.~Li, ``Lending orientation to neural networks for cross-view geo-localization,'' in \emph{Proceedings of the IEEE Conference on Computer Vision and Pattern Recognition}, 2019, pp. 5624--5633.

\bibitem{hu2018cvm}
S.~Hu, M.~Feng, R.~M. Nguyen, and G.~H. Lee, ``Cvm-net: Cross-view matching network for image-based ground-to-aerial geo-localization,'' in \emph{Proceedings of the IEEE Conference on Computer Vision and Pattern Recognition}, 2018, pp. 7258--7267.

\bibitem{arandjelovic2016netvlad}
R.~Arandjelovic, P.~Gronat, A.~Torii, T.~Pajdla, and J.~Sivic, ``Netvlad: Cnn architecture for weakly supervised place recognition,'' in \emph{Proceedings of the IEEE Conference on Computer Vision and Pattern Recognition}, 2016, pp. 5297--5307.

\bibitem{regmi2019bridging}
K.~Regmi and M.~Shah, ``Bridging the domain gap for ground-to-aerial image matching,'' in \emph{Proceedings of the IEEE/CVF International Conference on Computer Vision}, 2019, pp. 470--479.

\bibitem{zhang2023cross}
X.~Zhang, X.~Li, W.~Sultani, Y.~Zhou, and S.~Wshah, ``Cross-view geo-localization via learning disentangled geometric layout correspondence,'' in \emph{Proceedings of the AAAI Conference on Artificial Intelligence}, vol.~37, no.~3, 2023, pp. 3480--3488.

\bibitem{Deuser_2023_ICCV}
F.~Deuser, K.~Habel, and N.~Oswald, ``Sample4geo: Hard negative sampling for cross-view geo-localisation,'' in \emph{Proceedings of the IEEE/CVF International Conference on Computer Vision}, 2023, pp. 16\,847--16\,856.

\bibitem{caron2021emerging}
M.~Caron, H.~Touvron, I.~Misra, H.~J{\'e}gou, J.~Mairal, P.~Bojanowski, and A.~Joulin, ``Emerging properties in self-supervised vision transformers,'' in \emph{Proceedings of the IEEE/CVF International Conference on Computer Vision}, 2021, pp. 9650--9660.

\bibitem{oquab2024dinov}
M.~Oquab, T.~Darcet, T.~Moutakanni, H.~V. Vo, M.~Szafraniec, V.~Khalidov, P.~Fernandez \emph{et~al.}, ``{DINO}v2: Learning robust visual features without supervision,'' \emph{Transactions on Machine Learning Research}, 2024.

\bibitem{deng2009imagenet}
J.~Deng, W.~Dong, R.~Socher, L.-J. Li, K.~Li, and L.~Fei-Fei, ``Imagenet: A large-scale hierarchical image database,'' in \emph{Proceedings of the IEEE/CVF Conference on Computer Vision and Pattern Recognition}, 2009, pp. 248--255.

\bibitem{radford2021learning}
A.~Radford, J.~W. Kim, C.~Hallacy, A.~Ramesh, G.~Goh, S.~Agarwal, G.~Sastry, A.~Askell, P.~Mishkin, J.~Clark \emph{et~al.}, ``Learning transferable visual models from natural language supervision,'' in \emph{International Conference on Machine Learning}.\hskip 1em plus 0.5em minus 0.4em\relax PMLR, 2021, pp. 8748--8763.

\bibitem{kirillov2023segment}
A.~Kirillov, E.~Mintun, N.~Ravi, H.~Mao, C.~Rolland, L.~Gustafson, T.~Xiao, S.~Whitehead, A.~C. Berg, W.-Y. Lo \emph{et~al.}, ``Segment anything,'' \emph{arXiv preprint arXiv:2304.02643}, 2023.

\bibitem{wu2022im2city}
M.~Wu and Q.~Huang, ``Im2city: image geo-localization via multi-modal learning,'' in \emph{Proceedings of the 5th ACM SIGSPATIAL International Workshop on AI for Geographic Knowledge Discovery}, 2022, pp. 50--61.

\bibitem{cepeda2023geoclip}
V.~V. Cepeda, G.~K. Nayak, and M.~Shah, ``Geoclip: Clip-inspired alignment between locations and images for effective worldwide geo-localization,'' in \emph{Thirty-seventh Conference on Neural Information Processing Systems}, 2023.

\bibitem{haas2023learning}
L.~Haas, S.~Alberti, and M.~Skreta, ``Learning generalized zero-shot learners for open-domain image geolocalization,'' \emph{arXiv preprint arXiv:2302.00275}, 2023.

\bibitem{keetha2023anyloc}
N.~Keetha, A.~Mishra, J.~Karhade, K.~M. Jatavallabhula, S.~Scherer, M.~Krishna, and S.~Garg, ``Anyloc: Towards universal visual place recognition,'' \emph{IEEE Robotics and Automation Letters}, 2023.

\bibitem{xiao2023visual}
J.~Xiao, G.~Zhu, and G.~Loianno, ``Visual geo-localization with self-supervised representation learning,'' \emph{arXiv preprint arXiv:2308.00090}, 2023.

\bibitem{pantazis2022svl}
O.~Pantazis, G.~Brostow, K.~Jones, and O.~Mac~Aodha, ``Svl-adapter: Self-supervised adapter for vision-language pretrained models,'' in \emph{Proceedings of The 33rd British Machine Vision Conference}.\hskip 1em plus 0.5em minus 0.4em\relax The British Machine Vision Association, 2022.

\bibitem{pan2023self}
K.~Pan, J.~Li, H.~Song, J.~Lin, X.~Liu, and S.~Tang, ``Self-supervised meta-prompt learning with meta-gradient regularization for few-shot generalization,'' \emph{arXiv preprint arXiv:2303.12314}, 2023.

\bibitem{lu2023towards}
F.~Lu, L.~Zhang, X.~Lan, S.~Dong, Y.~Wang, and C.~Yuan, ``Towards seamless adaptation of pre-trained models for visual place recognition,'' in \emph{The Twelfth International Conference on Learning Representations}, 2023.

\bibitem{chen2023sam}
T.~Chen, L.~Zhu, C.~Deng, R.~Cao, Y.~Wang, S.~Zhang, Z.~Li, L.~Sun, Y.~Zang, and P.~Mao, ``Sam-adapter: Adapting segment anything in underperformed scenes,'' in \emph{Proceedings of the IEEE/CVF International Conference on Computer Vision}, 2023, pp. 3367--3375.

\bibitem{yang2020mining}
F.~Yang, Z.~Wang, J.~Xiao, and S.~Satoh, ``Mining on heterogeneous manifolds for zero-shot cross-modal image retrieval,'' in \emph{Proceedings of the AAAI Conference on Artificial Intelligence}, vol.~34, no.~07, 2020, pp. 12\,589--12\,596.

\bibitem{xu2022bayesian}
M.-C. Xu, Y.~Zhou, C.~Jin, M.~de~Groot, D.~C. Alexander, N.~P. Oxtoby, Y.~Hu, and J.~Jacob, ``Bayesian pseudo labels: Expectation maximization for robust and efficient semi-supervised segmentation,'' in \emph{International Conference on Medical Image Computing and Computer-Assisted Intervention}.\hskip 1em plus 0.5em minus 0.4em\relax Springer, 2022, pp. 580--590.

\bibitem{neal1998view}
R.~M. Neal and G.~E. Hinton, ``A view of the em algorithm that justifies incremental, sparse, and other variants,'' in \emph{Learning in graphical models}.\hskip 1em plus 0.5em minus 0.4em\relax Springer, 1998, pp. 355--368.

\bibitem{sumbul2022novel}
G.~Sumbul, M.~M{\"u}ller, and B.~Demir, ``A novel self-supervised cross-modal image retrieval method in remote sensing,'' in \emph{2022 IEEE International Conference on Image Processing}.\hskip 1em plus 0.5em minus 0.4em\relax IEEE, 2022, pp. 2426--2430.

\bibitem{wang2022multiple}
T.~Wang, Z.~Zheng, Y.~Sun, T.-S. Chua, Y.~Yang, and C.~Yan, ``Multiple-environment self-adaptive network for aerial-view geo-localization,'' \emph{arXiv preprint arXiv:2204.08381}, 2022.

\bibitem{bommasani2021opportunities}
R.~Bommasani, D.~A. Hudson, E.~Adeli, R.~Altman, S.~Arora, S.~von Arx, M.~S. Bernstein, J.~Bohg, A.~Bosselut, E.~Brunskill \emph{et~al.}, ``On the opportunities and risks of foundation models,'' \emph{arXiv preprint arXiv:2108.07258}, 2021.

\bibitem{caron2020unsupervised}
M.~Caron, I.~Misra, J.~Mairal, P.~Goyal, P.~Bojanowski, and A.~Joulin, ``Unsupervised learning of visual features by contrasting cluster assignments,'' \emph{Advances in Neural Information Processing Systems}, vol.~33, pp. 9912--9924, 2020.

\bibitem{oord2018representation}
A.~v.~d. Oord, Y.~Li, and O.~Vinyals, ``Representation learning with contrastive predictive coding,'' \emph{arXiv preprint arXiv:1807.03748}, 2018.

\bibitem{gao2023clip}
P.~Gao, S.~Geng, R.~Zhang, T.~Ma, R.~Fang, Y.~Zhang, H.~Li, and Y.~Qiao, ``Clip-adapter: Better vision-language models with feature adapters,'' \emph{International Journal of Computer Vision}, pp. 1--15, 2023.

\bibitem{zheng2023uavs}
Z.~Zheng, Y.~Shi, T.~Wang, J.~Liu, J.~Fang, Y.~Wei, and T.-s. Chua, ``Uavs in multimedia: Capturing the world from a new perspective,'' in \emph{Proceedings of the 31th ACM International Conference on Multimedia Workshop}, vol.~4, 2023.

\bibitem{he2016deep}
K.~He, X.~Zhang, S.~Ren, and J.~Sun, ``Deep residual learning for image recognition,'' in \emph{Proceedings of the IEEE Conference on Computer Vision and Pattern Recognition}, 2016, pp. 770--778.

\bibitem{ding2020practical}
L.~Ding, J.~Zhou, L.~Meng, and Z.~Long, ``A practical cross-view image matching method between uav and satellite for uav-based geo-localization,'' \emph{Remote Sensing}, vol.~13, no.~1, p.~47, 2020.

\bibitem{wang2022learning}
T.~Wang, Z.~Zheng, Z.~Zhu, Y.~Gao, Y.~Yang, and C.~Yan, ``Learning cross-view geo-localization embeddings via dynamic weighted decorrelation regularization,'' \emph{arXiv preprint arXiv:2211.05296}, 2022.

\bibitem{chen2022we}
C.~Chen, J.~Zhang, Y.~Xu, L.~Chen, J.~Duan, Y.~Chen, S.~Tran, B.~Zeng, and T.~Chilimbi, ``Why do we need large batchsizes in contrastive learning? a gradient-bias perspective,'' \emph{Advances in Neural Information Processing Systems}, vol.~35, pp. 33\,860--33\,875, 2022.

\end{thebibliography}
\end{document}